\documentclass[lettersize,journal]{IEEEtran}
\usepackage{amsmath,amsfonts}
\usepackage{algorithmic}
\usepackage{algorithm}
\usepackage{array}
\usepackage[caption=false,font=normalsize,labelfont=sf,textfont=sf]{subfig}
\usepackage{textcomp}
\usepackage{stfloats}
\usepackage{url}
\usepackage{verbatim}
\usepackage{graphicx}
\usepackage{cite}
\usepackage{newtxmath}
\usepackage{multirow}
\usepackage{booktabs}
\usepackage{CJK}
\usepackage{xpinyin}
\usepackage{color, xcolor}

\begin{document}

\title{On the (In)Effectiveness of Large Language Models\\ for Chinese Text Correction}

\author{Yinghui Li, Haojing Huang, Shirong Ma, Yong Jiang, Yangning Li, \\
Feng Zhou, Hai-Tao Zheng~\IEEEmembership{Member,~IEEE}, Qingyu Zhou
% <-this % stops a space
\thanks{Manuscript received Dec 11, 2023. \textit{( Yinghui Li, Haojing Huang, and Shirong Ma contributed equally to this work. Corresponding authors: Hai-Tao Zheng and Qingyu Zhou.)}}%
\thanks{Yinghui Li, Haojing Huang, Shirong Ma, Yangning Li, and Hai-Tao Zheng are with Tsinghua Shenzhen International Graduate School, Tsinghua University, Shenzhen, Guangdong 518055, China. E-mail: liyinghu20@mails.tsinghua.edu.cn,  haojing.huangg@gmail.com, masr21@mails.tsinghua.edu.cn, yn-li23@mails.tsinghua.edu.cn, zheng.haitao@sz.tsinghua.edu.cn.}
\thanks{Yong Jiang is with DAMO Academy, Alibaba Group, Hangzhou, Zhejiang 311100, China. E-mail: jiangyong.ml@gmail.com.}
\thanks{Feng Zhou and Qingyu Zhou are with OPPO Research Institute, Beijing 100026, China. E-mail: zhoufeng1@oppo.com, qyzhgm@gmail.com.}
}

\markboth{Journal of \LaTeX\ Class Files,~Vol.~14, No.~8, August~2021}%
{Shell \MakeLowercase{\textit{et al.}}: A Sample Article Using IEEEtran.cls for IEEE Journals}

\IEEEpubid{0000--0000/00\$00.00~\copyright~2021 IEEE}

\maketitle
\begin{CJK*}{UTF8}{gbsn}
\begin{abstract}
  Recently, the development and progress of Large Language Models (LLMs) have amazed the entire Artificial Intelligence community. Benefiting from their emergent abilities, LLMs have attracted more and more researchers to study their capabilities and performance on various downstream Natural Language Processing (NLP) tasks. While marveling at LLMs' incredible performance on all kinds of tasks, we notice that they also have excellent multilingual processing capabilities, such as Chinese. To explore the Chinese processing ability of LLMs, we focus on Chinese Text Correction, a fundamental and challenging Chinese NLP task. Specifically, we evaluate various representative LLMs on the Chinese Grammatical Error Correction (CGEC) and Chinese Spelling Check (CSC) tasks, which are two main Chinese Text Correction scenarios. Additionally, we also fine-tune LLMs for Chinese Text Correction to better observe the potential capabilities of LLMs. From extensive analyses and comparisons with previous state-of-the-art small models, we empirically find that the LLMs currently have both amazing performance and unsatisfactory behavior for Chinese Text Correction. We believe our findings will promote the landing and application of LLMs in the Chinese NLP community.
\end{abstract}


\section{Introduction}
\IEEEPARstart{L}{arge} Language Models (LLMs) have achieved remarkable progress in the last few months and are gradually becoming the fundamental infrastructure in the field of Natural Language Processing (NLP)~\cite{zhao2023survey}. 
Benefiting from the emergent abilities~\cite{wei2022emergent} and the advantages of the chain-of-thought~\cite{NEURIPS2022_9d560961}, LLMs seem to be sweeping various downstream tasks of NLP with a unified conversational paradigm. In the past few months, LLMs have been extensively evaluated on all kinds of NLP tasks and have shown performance beyond expectations, such as Natural Language Understanding~\cite{he2023chatgpt}, Information Extraction~\cite{wei2023zeroshot}, and Text Summarization~\cite{yang2023exploring}. In addition to LLMs' general ability for various tasks, their excellent multilingual adaptability is also impressive. As one of the most spoken languages in the world, Chinese has always been a challenging and research-worthy language in the NLP community because of its unique language nature and characteristics~\cite{liu-etal-2010-visually}. Therefore, this paper conducts a comprehensive evaluation of LLMs' capability on a fundamental and challenging Chinese NLP task: Chinese Text Correction~\cite{zhao2022overview}.

Chinese Text Correction aims to detect and correct various errors contained in the input Chinese text~\cite{zhao2022overview}. According to the types of errors, Chinese Text Correction is generally divided into two categories of tasks: Chinese Grammatical Error Correction (CGEC)~\cite{wang2020comprehensive, ye2022focus} and Chinese Spelling Check (CSC)~\cite{wu-etal-2013-integrating, li-etal-2022-past, zhang2023contextual}.
The CGEC task focuses on the grammatical errors that Chinese-as-a-Second-Language (CSL) learners tend to make during their language learning process and Chinese native speakers accidentally make in their daily lives. Due to the difference in language proficiency, the focus of the CGEC task for CSL learners and the CGEC task for Chinese native speakers is also different. Because CSL learners do not have a high level of mastery of Chinese, the grammatical errors they often make mainly involve the addition, deletion, replacement, and reordering of characters, while the grammatical errors made by Chinese native speakers are often more subtle and difficult, which has higher requirements and challenges for the model to understand the Chinese grammatical rules. 
As for the CSC task, it is for automatically checking spelling errors in Chinese text. Due to the characteristics of Chinese characters, Chinese spelling errors are mainly caused by phonetically or visually similar characters. Therefore, CSC is challenging because it requires not only complex semantic knowledge, but also phonetics/vision information to assist the models in finding the correct characters. It can be seen from the above that \textbf{Chinese Text Correction is a practical and complex Chinese application scenario, and studying the Chinese Text Correction ability of LLMs well reflects their Chinese processing ability.}

\IEEEpubidadjcol
In this paper, we conduct a comprehensive study to evaluate the Chinese error correction ability of LLMs. First, according to the characteristics of CGEC and CSC, we carefully design task-specific prompts to guide LLMs to behave like a corrector. Then, we explore some widely used in-context learning prompting strategies to further inspire the correction ability of LLMs. In addition, we also construct instruction training data based on existing Chinese Text Correction training data to fine-tune LLMs, thus further enhancing their correction ability.
We conduct extensive experiments on CSL and native CGEC benchmarks, as well as CSC datasets from multiple domains. It is worth noting that the charm of text correction is that it is an extremely subjective task, that is, there may be multiple reference sentences for an erroneous sentence, so the automatic evaluation of benchmarks and objective metrics may not truly reflect the performance of the model, hence, we also conduct a deep manual evaluation to observe LLMs more realistic correction ability.

Through comparisons with previous state-of-the-art fine-tuned models and detailed analyses, we obtain the following findings and insights:
\begin{itemize}
    \item Although there is a significant difference between the automatic and human evaluation results, Chinese Text Correction is still very challenging for LLMs. And the performance of LLMs still has a large gap with the previous fine-tuned small models. For Chinese Text Correction, LLMs also have some bright spots. We find that LLMs have better domain adaptability and data tolerance ability than traditional models.
    \item Well-designed prompts and in-context learning strategies effectively improve the Chinese Text Correction ability of LLMs. Therefore, for Chinese Text Correction, when designing prompts and in-context example selection strategies, the characteristics and settings of the Chinese Text Correction tasks must be considered.
    \item Different base model selections and different fine-tuning strategies have different impacts on Chinese Text Correction. Additionally, the construction of instruction data for fine-tuning LLMs is also critical to their performance on Chinese Text Correction tasks.
\end{itemize}

\begin{figure*}
\centering
\includegraphics[width=1.00\textwidth]{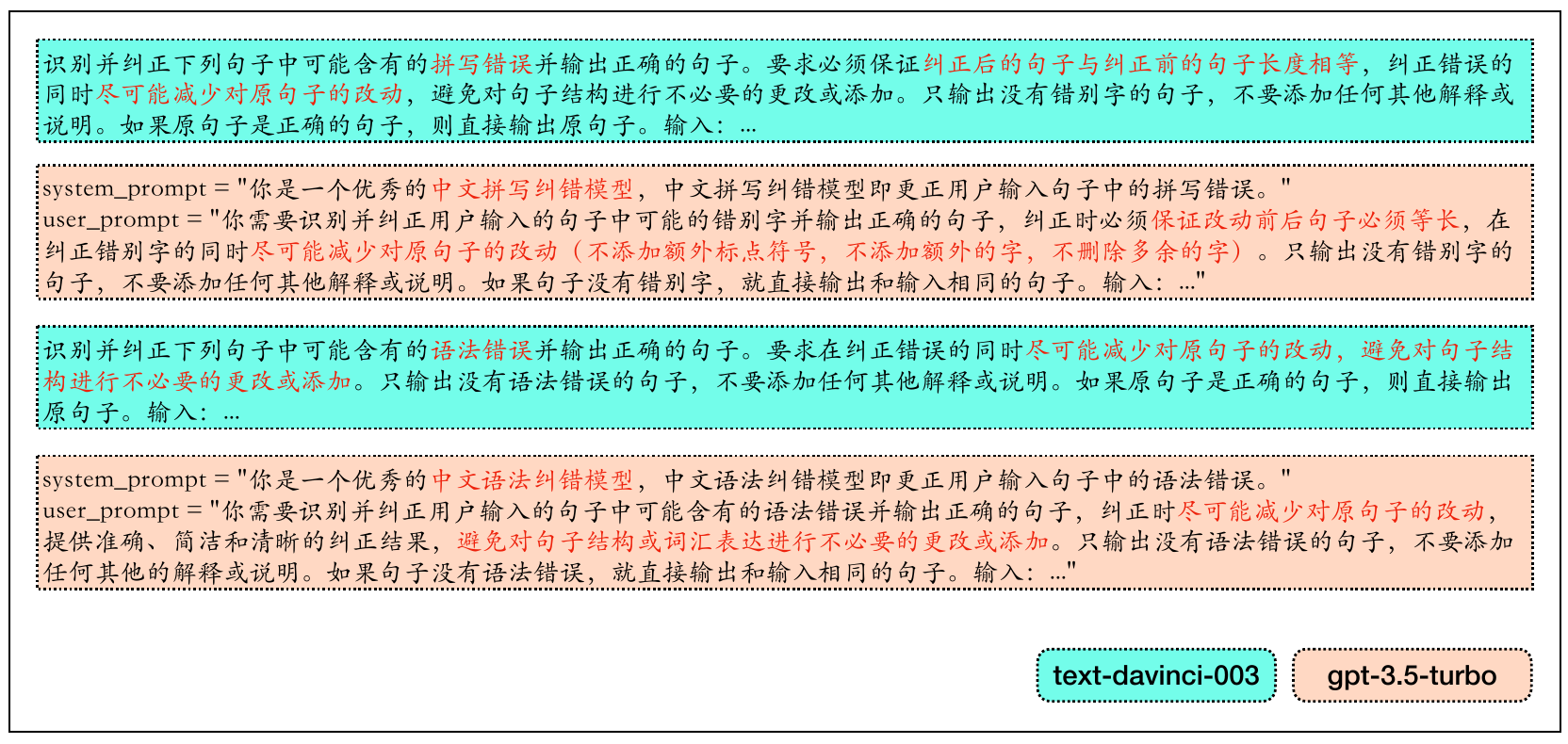}
\caption{Task-specific prompts of the CSC (中文拼写纠错) and CGEC (中文语法纠错) tasks. In our study, we try different ChatGPT base models, such as \texttt{text-davinci-003} and \texttt{gpt-3.5-turbo}, and other LLMs. We mark the key information related to the task characteristics in the prompt in red.}
\label{Figure:Prompt}
\end{figure*}

\section{Methodology}
\subsection{Task-specific Prompts}
To guide LLMs to behave like a CGEC model or a CSC model, we manually design task-specific prompts as shown in Figure~\ref{Figure:Prompt}. 
In fact, LLMs' general-purpose power makes them have a certain text polishing ability, but we found that if there are no task-specific restrictions on the input prompts, LLMs are very easy to play freely when doing text polishing or error correction, thus violating some basic settings and principles of the Chinese Text Correction task.
Therefore, considering that most Chinese Text Correction datasets and evaluation metrics focus on the minimum change principle~\cite{nagata-sakaguchi-2016-phrase}, that is, the model is required to make as few edit operations as possible to the input sentence, \textbf{we ask LLMs to minimize the changes to the original input sentence in the prompt}. In addition, for the CSC sub-task, since its task setting is that the input and output sentences are of equal length, \textbf{we require LLMs to ensure that the length of the corrected sentence is equal to the length of the sentence before correction, to avoid adding or deleting punctuation marks or Chinese characters}. As for the CGEC sub-task, because many grammatical errors involve making changes to the sentence structure, such as structural confusion and improper word order, to avoid some meaningless and unnecessary edits made by LLMs, \textbf{we also ask it to avoid unnecessary edits to the sentence structure or word expression}.

It is worth mentioning that the task-specific constraints we add in the prompts do not imply that LLMs cannot perform text correction without these constraints. Rather, if LLMs are to generate sentences without these constraints, they could have more creative freedom, which could create a gap between the grammatically correct sentences they generate and the existing Chinese Text Correction datasets and evaluation metrics. This gap may result in lower performance in error correction when evaluated. Therefore, to objectively and realistically evaluate LLMs' error correction performance on existing traditional datasets, we carefully designed the prompts in Figure~\ref{Figure:Prompt}.

\subsection{In-context Learning Strategies}
Many works~\cite{xie2022an,bansal-etal-2023-rethinking,dai-etal-2023-gpt} have shown that LLMs possess extraordinary in-context learning ability~\cite{dong2023survey}, i.e., by giving LLMs a small number of task examples to enhance their performance on specific tasks. 
To comprehensively study the in-context learning ability of LLMs in Chinese Text Correction, we design the following three sample selection strategies: 
\begin{enumerate}
% 普通的few shot，随机选取数据集中几条需要改正的sample，用“例如：xxx应该改为xxx”这样的方式加入prompt中，实验了1shot、2shot、5shot的性能
    \item \textbf{Select random erroneous samples}: Randomly select several sentences with text errors and their corresponding correct sentences from the dataset.
% 在few shot中加入正确样例，这种做法能让LLM对输出的格式更贴合我们要求的形式，也更能理解任务，例如2 shot中会包含1个需要修改的样例，一个不需要修改的样例，5shot包含三个需要修改的样例，2个不需要修改的样例
    \item \textbf{Select correct and erroneous samples}: Both samples without text errors and samples with text errors are selected. The purpose of this design is to teach LLMs to correct mistakes while not forgetting to modify the correct input sentences.
% similar shot，使用排序查找算法在数据集中找到与输入句子最相似的几个样例，作为few shot的例子加入到prompt中。
    \item \textbf{Select hard erroneous samples}: From the dataset, select the confusing samples where the wrong sentence is similar to the correct sentence, that is, these samples are challenging for the model to correct errors. 
    Specifically, we employ the BM25 algorithm to retrieve $N$ samples ($N=100$ in our work) from specified training datasets that are similar to the input sentence, including source and target sentences. Then, these samples are ranked based on the ROUGE-L similarity between the source sentence and the input sentence. We select the Top-$K$ samples with the highest ROUGE-L scores and integrate them into the prompt.
    We hope that by introducing such hard samples, LLMs can be stimulated to handle complex text errors.
\end{enumerate}

\subsection{Supervised Instruction Tuning}
\noindent\textbf{Collecting Raw Data:} For the CSC task, we select Wang271K~\cite{wang-etal-2018-hybrid} and SIGHAN~\cite{wu-etal-2013-sighan2013,yu-etal-2014-sighan2014,tseng-etal-2015-sighan2015} training data as the input-output pair raw data. Wang271K is a generated pseudo dataset consisting of 271,329 training examples. The SIGHAN13, SIGHAN14, and SIGHAN15 training examples number 700, 3437, and 2338, respectively. Originally, the SIGHAN datasets are in the Traditional Chinese. Following previous works~\cite{xu-etal-2021-read}, we convert them to Simplified Chinese using the OpenCC\footnote{\url{https://github.com/BYVoid/OpenCC}} tool.
For the CGEC task, we select data of FCGEC~\cite{xu-etal-2022-fcgec} as the input-output pair raw data. FCGEC is a human-annotated corpus of CGEC, comprised of sentences from public school Chinese examinations. Notably, all errors in the corpus were made by native speakers. There are 36340 training examples of FCGEC.

\noindent\textbf{Mapping Raw Data to Instruction Data:} Leveraging the collected raw data, we refine the instruction data. We begin by formulating the structure of the instruction data, incorporating four main elements:
1. Instruction of the task. We asked domain experts to write the instructions for CSC and CGEC. There are ten different instructions for each task. 
2. Example input-output pair (Optional). Inspired by the capability of In-context Learning of LLM, we try to add some data with few-shot examples into the instruction data. 
3. Input text. The sentence may contain spelling errors or grammatical errors. 
4. Output text: The corrected sentence.

We also observed that the performance of instruction tuning can be enhanced by incorporating universal instructions~\cite{DBLP:journals/corr/abs-2305-13225} Thus, we include general data including a wide range of questions asking LLM to solve. In practice, we use the \textit{alpaca\_gpt4\_data\_zh} provided in the \textit{GPT-4-LLM} project\footnote{\url{https://github.com/Instruction-Tuning-with-GPT-4/GPT-4-LLM}}.

\noindent\textbf{Instruction Tuning:} Based on the performance of LLMs in CTC, we select \texttt{Baichuan-13B-Chat} as our base model for the fine-tuning experiments. We use two different fine-tuning strategies, full-parameters fine-tuning (FT) and LoRA, following the \textit{LLaMA-Efficient-Tuning} project\footnote{\url{https://github.com/hiyouga/LLaMA-Efficient-Tuning}}. LoRA is a Parameter-Efficient Fine-Tuning technology, which freezes the pre-trained model weights and injects trainable rank decomposition matrices into each layer of the Transformer architecture. On the contrary, the full-parameters fine-tuning retrains all parameters of the base model.

\section{Experiments}
In this section, we present our experimental setup and findings on two Chinese Text Correction tasks, namely CSC and CGEC.

% Dataset
\begin{table}[h]
\small
\centering
\begin{tabular}{lrrr}
\toprule CSC Test Data & \#Sent & Avg. Length & \#Errors \\
\hline SIGHAN15 & 1,100 & 30.6 & 703 \\
LAW & 500 & 29.7 & 390 \\
MED & 500 & 49.6 & 356 \\
ODW & 500 & 40.5 & 404 \\
MCSCSet$^\dag$ & 1,000 & 10.9 & 919 \\
% \hline Total & 277,804 & 42.6 & 390464 \\
\hline \hline CGEC Test Data & \#Sent & Avg. Length & \#Errors \\
\hline NLPCC & 2,000 & 29.7 & 1,981 \\
MuCGEC & 1,137 & 44.0 & 1,082 \\
NaCGEC (Dev.) & 500 & 56.2 & 482 \\
% \hline Total & 3,162 & 50.9 & 2,698 \\
\bottomrule
\end{tabular}

\caption{Statistics of the datasets that we use in experiments. We report the number of sentences (\#Sent), the average sentence length (Avg. Length), and the number of spelling errors (\#Errors). $\dag$ means we randomly sample 1,000 items from the raw 19,650 test set.}
\label{Data_Statistics}
\end{table}

\subsection{Experimental Settings}
\noindent\textbf{Test Data:} For the CSC task, we select three widely used datasets for evaluation. \textbf{SIGHAN15}~\cite{tseng-etal-2015-sighan2015}, which is written in hands by Chinese-as-a-Second Language (CSL) learners. A domain-specific dataset~\cite{Lv_2023}, which includes three domains. \textbf{LAW} data is collected from question stems and options of multiple-choice questions in the judicial examination. \textbf{MED} data is collected from QA pairs from online medical treatment consultations. \textbf{ODW} is official document writing data comprising of news, policies, and national conditions officially reported by the state.  A medical-domain dataset \textbf{MCSCSet}~\cite{10.1145/3511808.3557636}, which is collected from extensive real-world medical queries from Tecent Yidian. For the CGEC task, we use three widely used datasets for evaluation. \textbf{NLPCC}~\cite{10.1007/978-3-319-99501-4_41} is the GEC task in the NLPCC 2018 shared tasks. \textbf{MuCGEC}~\cite{zhang-etal-2022-mucgec} is a multi-reference multi-source evaluation dataset collected from CSL learner sources. \textbf{NaCGEC}~\cite{ma-etal-2022-linguistic} is a dataset that the grammatical errors made by native Chinese speakers in real-world scenarios, such as examinations and news sites.
The statistics of the used datasets are shown in Table \ref{Data_Statistics}.

\noindent\textbf{Evaluation Metrics:} 
For the CSC task, we adopt the precision, recall, and F1 scores on sentence level as the evaluation metrics following the previous work~\cite{xu-etal-2021-read}. Sentence-level metrics are stricter than character-level metrics since a sentence is considered to be correct if and only if all errors in the sentence are successfully detected and corrected. The main results are reported on the detection and correction sub-tasks. 
For the CGEC task, we employ a character-based span-level ChERRANT scorer~\cite{zhang-etal-2022-mucgec}
for evaluation, which computes Precision, Recall, and F$_{0.5}$ between the gold edit set and the system edit set.

% Incorporating human evaluation allows for a more holistic and comprehensive assessment of the LLM's abilities. It provides a means to evaluate subjective aspects, such as content relevance, coherence, fluency, and the overall quality of generated output. By including human evaluators, we can consider the contextual and semantic aspects that are difficult to capture solely through automated metrics.

% Furthermore, human evaluation plays a vital role in addressing the ethical and societal implications of LLMs. These models have the potential to impact public discourse and decision-making processes. Human evaluators can help identify biases, potential misinformation, or outputs that may be harmful or misleading. By integrating human evaluation, we ensure a responsible and ethical use of LLM technology.

% In the following sections, we present our approach to human evaluation, including the guidelines provided to evaluators, the evaluation criteria employed, and the process for obtaining reliable and consistent assessments. Through this human evaluation component, we aim to provide a more comprehensive understanding of the LLM's performance and its potential implications.

\noindent\textbf{Compared Methods:} 
To evaluate the performance of LLMs, we select some advanced and strong CSC and CGEC models as our baselines: 
\textbf{BERT}~\cite{devlin-etal-2019-bert} encodes the input sentence first. Then, a classifier (e.g., linear layer) is used to pick the correction character from the vocabulary for each character. 
\textbf{SoftMasked-BERT}~\cite{zhang-etal-2020-spelling} is consist of Detection Network and Correction Network.
\textbf{MedBERT-Corrector}~\cite{10.1145/3511808.3557636} is a CSC model that takes advantage of the domain knowledge in medicine.
\textbf{REALISE}~\cite{xu-etal-2021-read} is a multimodal CSC model that leverages semantic, phonetic, and graphic knowledge.
\textbf{Two-Ways}~\cite{li-etal-2021-exploration} utilizes the weak spots of the model to generate pseudo-training data.
\textbf{LEAD}~\cite{li-etal-2022-learning-dictionary} learns heterogeneous knowledge from the dictionary, especially the knowledge of definition. 
\textbf{BART-Large-Chinese}~\cite{shao2021cpt} is a Chinese BART model employed as a sequence-to-sequence (Seq2Seq) grammatical error correction baseline. BART~\cite{lewis2020bart} is a Transformer-based Seq2Seq pre-trained model~\cite{dong2022survey}, which utilizes denoising auto-encoder (DAE) as the pre-training task.
\textbf{GECToR-Chinese}~\cite{zhang2022mucgec} is a Chinese variant of GECToR using StructBERT~\cite{wei2020structbert} as its semantic encoder. GECToR~\cite{omelianchuk2020gector} is an iterative sequence-to-edit (Seq2Edit) GEC approach.

For the selection of LLMs, OpenAI provides a range of LLMs, which are accessible through public APIs\footnote{\url{https://chat.openai.com/}}. In our research, we specifically choose two advanced models from the \textbf{ChatGPT} series: \texttt{text-davinci-003} and \texttt{gpt-3.5-turbo}, as the focus of our investigation. 
The \texttt{text-davinci-003} is a variant of GPT-3 that has been trained using Reinforcement Learning from Human Feedback (RLHF). On the other hand, \texttt{gpt-3.5-turbo} serves as the underlying model for ChatGPT, incorporating a larger dataset and also trained on GPT-3 with RLHF. In addition, because our research is mainly focused on Chinese scenarios, we also select well-known and widely used LLMs in the Chinese NLP community for research, such as Vicuna~\cite{vicuna2023}, ChatGLM~\cite{du2022glm}, ChatGLM2~\cite{du2022glm}, and Baichuan\footnote{\url{https://github.com/baichuan-inc/Baichuan-13B}}. Notably, the ChatGLM series of models only open source the parameters of the 6B and 130B scales. Therefore, for a fair comparison with other LLMs, we choose the 6B models of the ChatGLM series.

% Main Result CSC
\begin{table*}[ht]
\footnotesize
\centering
\begin{tabular}{@{}c|l|ccc|ccc@{}}
\toprule
\multicolumn{1}{c|}{\textbf{Dataset}} & \multicolumn{1}{c|}{\textbf{Model}}                & \multicolumn{3}{c|}{\textbf{Detection}}          & \multicolumn{3}{c}{\textbf{Correction}}          \\
                             &                      & \textbf{P}           & \textbf{R}           & \textbf{F}           & \textbf{P}           & \textbf{R}           & \textbf{F}           \\ 
                             \midrule
\multicolumn{1}{c|}{\multirow{9}{*}{\textbf{SIGHAN15}}}   & REALISE~\cite{xu-etal-2021-read}          & 77.3 & 81.3          & 79.3          & 75.9          & 79.9          & 77.8          \\
\multicolumn{1}{c|}{\multirow{9}{*}{~\cite{tseng-etal-2015-sighan2015}}} & Two-Ways~\cite{li-etal-2021-exploration}        & -           & -             & 80.0          & -           & -           & 78.2          \\
\multicolumn{1}{c|}{}                             & LEAD~\cite{li-etal-2022-learning-dictionary}                 & 79.2          & 82.8          & 80.9          & 77.6          & 81.2          & 79.3          \\  
                                        \cmidrule(l){2-8} 
\multicolumn{1}{l|}{}                             & text-davinci-003              & 21.2          & 36.8          & 26.9          & 15.4          & 26.6          & 19.5          \\
\multicolumn{1}{l|}{}                             & gpt-3.5-turbo          & 17.2  & 30.1  & 21.9  & 14.3  & 25.1  & 18.2  \\

\multicolumn{1}{l|}{}  & Vicuna-13B-v1.3     & 6.3 & 7.9  & 7.0  & 2.6 & 3.3  & 2.9 \\
\multicolumn{1}{l|}{}  & ChatGLM-6B      & 2.7  & 4.1  & 3.3  & 1.4  & 2.0  & 1.6   \\
\multicolumn{1}{l|}{}  & ChatGLM2-6B      & 4.9 & 7.8  & 6.0  & 2.4 & 3.9  & 3.0 \\
\multicolumn{1}{l|}{}  & Baichuan-13B-Chat  & 8.3 & 14.0 & 10.4 & 6.1 & 10.4 & 7.7 \\
%                                     \cmidrule(l){2-8}
% % \multicolumn{1}{l|}{}  & ChatGLM-6B (LoRA)      & 30.0 & 28.7 & 29.3 & 18.6 & 17.7 & 18.2 \\
% \multicolumn{1}{l|}{}  & Baichuan-13B-Chat (LoRA)  & 27.4 & 35.5 & 30.9 & 22.7 & 29.4 & 25.6 \\
                                        \midrule
\multicolumn{1}{c|}{\multirow{9}{*}{\textbf{LAW}}}  & Soft-Masked BERT~\cite{zhang-etal-2020-spelling}              & 55.2          & 49.3          & 52.1          & 39.8          & 35.6          & 37.6          \\
\multicolumn{1}{c|}{\multirow{9}{*}{~\cite{Lv_2023}}} & BERT~\cite{devlin-etal-2019-bert}              & 79.0          & 68.6          & 73.4          & 71.4          & 62.0          & 66.3          \\
\multicolumn{1}{l|}{}                               & ECSpell~\cite{Lv_2023}              & 78.2          & 67.8          & 72.6          & 72.2          & 62.6          & 67.1         \\
                                        \cmidrule(l){2-8}
\multicolumn{1}{l|}{}                             & text-davinci-003              & 27.8          & 42.0          & 33.4          & 23.6          & 35.7          & 28.4          \\
\multicolumn{1}{l|}{}                             & gpt-3.5-turbo           &42.0    & 43.5   & 42.8   & 33.7   & 34.9   & 34.3  \\ 

\multicolumn{1}{l|}{}  & Vicuna-13B-v1.3     & 4.8  & 5.1  & 4.9  & 2.2  & 2.4  & 2.3  \\
\multicolumn{1}{l|}{}  & ChatGLM-6B      & 6.8  & 8.2  & 7.5  & 5.5  & 6.7  & 6.0  \\
\multicolumn{1}{l|}{}  & ChatGLM2-6B       & 13.0 & 13.3 & 13.2 & 10.3 & 10.6 & 10.4 \\
\multicolumn{1}{l|}{}  & Baichuan-13B-Chat  & 21.6 & 32.9 & 26.1 & 18.0 & 27.5 & 21.8 \\
%                                     \cmidrule(l){2-8}
% % \multicolumn{1}{l|}{}  & ChatGLM-6B (LoRA)      & 30.6 & 33.3 & 31.9 & 20.1 & 22.0 & 21.0 \\
% \multicolumn{1}{l|}{}  & Baichuan-13B-Chat (LoRA)   & 51.2 & 56.5 & 53.7 & 43.4 & 47.8 & 45.5 \\
                                        \midrule
\multicolumn{1}{c|}{\multirow{9}{*}{\textbf{MED}}}  & Soft-Masked BERT~\cite{zhang-etal-2020-spelling}              & 44.4          & 45.1          & 44.7          & 26.8          & 29.2          & 28.0          \\
\multicolumn{1}{c|}{\multirow{9}{*}{~\cite{Lv_2023}}}      & BERT~\cite{devlin-etal-2019-bert}              & 76.0          & 67.2          & 71.3          & 66.4          & 57.6          & 61.7         \\
\multicolumn{1}{l|}{}                               & ECSpell~\cite{Lv_2023}              & 75.8          & 65.8          & 70.4          & 67.6          & 58.6          & 62.8          \\
                                        \cmidrule(l){2-8} 
\multicolumn{1}{c|}{}                             & text-davinci-003                & 14.8          & 27.4          & 19.2          & 11.2          & 20.8          & 14.6          \\
\multicolumn{1}{l|}{}                             & gpt-3.5-turbo                     &30.5                      & 42.0                &35.4           &23.5               &32.3       &27.2   \\ 

\multicolumn{1}{l|}{}  & Vicuna-13B-v1.3     & 2.4  & 3.1  & 2.7  & 1.7  & 2.2  & 1.9  \\
\multicolumn{1}{l|}{}  & ChatGLM-6B     & 1.5  & 2.7  & 1.9  & 1.3  & 2.2  & 1.6  \\
\multicolumn{1}{l|}{}  & ChatGLM2-6B       & 7.0  & 10.2 & 8.3  & 6.1  & 8.8  & 7.2  \\
\multicolumn{1}{l|}{}  & Baichuan-13B-Chat  & 13.7 & 23.9 & 17.4 & 10.9 & 19.0 & 13.9 \\
%                                     \cmidrule(l){2-8}
% % \multicolumn{1}{l|}{}  & ChatGLM-6B (LoRA)     & 33.9 & 35.4 & 34.6 & 20.3 & 21.2 & 20.8 \\
% \multicolumn{1}{l|}{}  & Baichuan-13B-Chat (LoRA)   & 44.9 & 46.5 & 45.7 & 34.2 & 35.4 & 34.8 \\
                                        \midrule
\multicolumn{1}{c|}{\multirow{9}{*}{\textbf{ODW}}}  & Soft-Masked BERT~\cite{zhang-etal-2020-spelling}              & 53.4          & 49.0          & 51.1          & 37.5          & 35.1          & 36.3          \\
\multicolumn{1}{c|}{\multirow{9}{*}{~\cite{Lv_2023}}}   & BERT~\cite{devlin-etal-2019-bert}              & 82.6          & 65.8          & 73.2         & 75.9          & 62.1          & 68.3          \\ 
\multicolumn{1}{c|}{}             & ECSpell~\cite{Lv_2023}              & 82.4          & 70.1          & 75.8          & 76.9          & 62.1          & 70.0          \\
                                        \cmidrule(l){2-8}
\multicolumn{1}{c|}{}                             & text-davinci-003              & 38.0          & 50.8          & 43.5          & 30.9          & 41.2          & 35.3          \\
\multicolumn{1}{l|}{}                             & gpt-3.5-turbo          &63.8      & 54.6       & 58.8       & 52.2               & 44.7                     & 48.1               \\ 

\multicolumn{1}{l|}{}  & Vicuna-13B-v1.3     & 7.1  & 6.5  & 6.8  & 4.1  & 3.8  & 4.0  \\
\multicolumn{1}{l|}{}  & ChatGLM-6B     & 5.6  & 6.5  & 6.0  & 4.9  & 5.7  & 5.3  \\
\multicolumn{1}{l|}{}  & ChatGLM2-6B       & 15.6 & 16.4 & 16.0 & 13.8 & 14.5 & 14.2 \\
\multicolumn{1}{l|}{}  & Baichuan-13B-Chat  & 32.1 & 41.6 & 36.2 & 25.6 & 33.2 & 28.9 \\
%                                     \cmidrule(l){2-8}
% % \multicolumn{1}{l|}{}  & ChatGLM-6B (LoRA)     & 42.2 & 35.1 & 38.3 & 29.8 & 24.8 & 27.1 \\
% \multicolumn{1}{l|}{}  & Baichuan-13B-Chat (LoRA)   & 49.8 & 51.9 & 50.8 & 41.4 & 43.1 & 42.2 \\
                                        \midrule
\multicolumn{1}{c|}{\multirow{9}{*}{\textbf{MCSCSet}}} & BERT~\cite{devlin-etal-2019-bert}         & 87.1          & 86.1          & 86.6          & 80.9          & 80.1          & 80.5         \\
\multicolumn{1}{c|}{\multirow{9}{*}{~\cite{10.1145/3511808.3557636}}}     & MedBERT-Corrector~\cite{10.1145/3511808.3557636}        & 87.0          & 86.3          & 86.6          & 81.0          & 80.5         & 80.6          \\
\multicolumn{1}{l|}{}                             & Soft-Masked BERT~\cite{zhang-etal-2020-spelling}         & 87.0          & 86.3          & 86.7          & 81.2          & 80.5          & 80.9          \\ 
                                        \cmidrule(l){2-8}
\multicolumn{1}{l|}{}                             & text-davinci-003                & 23.9          & 36.4          & 28.8          & 12.5          & 19.0 & 15.0          \\
\multicolumn{1}{l|}{}                             & gpt-3.5-turbo            & 36.0     & 36.2    & 36.1      & 25.0          & 25.1            & 25.0                  \\  

\multicolumn{1}{l|}{}  & Vicuna-13B-v1.3     & 2.4  & 3.2  & 2.8  & 0.6  & 0.8  & 0.7  \\
\multicolumn{1}{l|}{}  & ChatGLM-6B     & 0.4  & 0.8  & 0.5  & 0.0  & 0.0  & 0.0  \\
\multicolumn{1}{l|}{}  & ChatGLM2-6B       & 3.5  & 5.0  & 4.1  & 2.3  & 3.4  & 2.8  \\
\multicolumn{1}{l|}{}  & Baichuan-13B-Chat  & 19.6 & 27.5 & 22.9 & 14.3 & 20.0 & 16.7 \\
%                                     \cmidrule(l){2-8}
% % \multicolumn{1}{l|}{}  & ChatGLM-6B (LoRA)     & 26.5 & 28.1 & 27.3 & 13.8 & 14.7 & 14.2 \\
% \multicolumn{1}{l|}{}  & Baichuan-13B-Chat (LoRA)   & 46.1 & 54.3 & 49.8 & 37.5 & 44.2 & 40.5 \\
                                        \bottomrule
\end{tabular}
\caption{
The automatically evaluated performance of LLMs and all baselines in CSC.}
\label{Table: Main_Results_CSC}
\end{table*}

% Main Results CGEC
\begin{table*}[ht]
\small
\centering
\begin{tabular}{@{}l|ccc|ccc|ccc@{}}
\toprule
\multirow{2}{*}{\textbf{MODEL}}                & \multicolumn{3}{c|}{\textbf{NLPCC}}          & \multicolumn{3}{c|}{\textbf{MuCGEC}}  & \multicolumn{3}{c}{\textbf{NaCGEC}}         \\
                                    & \textbf{P}       & \textbf{R}      & \textbf{F}       & \textbf{P}        & \textbf{R}       & \textbf{F}   & \textbf{P}       & \textbf{R}       & \textbf{F}      \\ 
                             \midrule

Seq2Seq-Baseline (BART-Large-Chinese)      & 37.0 & 26.1 & 34.1 & 38.9 & 28.0 & 36.1 & 19.0 & 7.6  & 14.6 \\
Seq2Edit-Baseline (GECToR-Chinese) & 42.5 & 24.7 & 37.2 & 36.5 & 27.4 & 34.2 & 14.7 & 15.3 & 14.9 \\

                                        \cmidrule(l){1-10} 
 text-davinci-003              & 19.6                & 23.1                & 20.2                & 29.3               & 26.0               & 28.6                & 6.7                 & 10.7                & 7.2                 \\
gpt-3.5-turbo          & 14.4                & 27.2                & 15.9                & 20.4                 & 32.4                & 22.0                & 5.6                 & 11.7                & 6.3                 \\

Vicuna-13B-v1.3     & 10.0 & 9.8  & 10.0 & 15.8    & 12.3    & 14.9    & 2.6  & 4.0  & 2.8  \\
ChatGLM-6B     & 6.0  & 10.8 & 6.5  & 8.9  & 12.3 & 9.4  & 2.5  & 7.2  & 2.9  \\
ChatGLM2-6B       & 10.5 & 10.6 & 10.5 & 13.3 & 12.7 & 13.2 & 3.3  & 6.7  & 3.6  \\
Baichuan-13B-Chat  & 6.3  & 18.4 & 7.3  & 8.3  & 2.2  & 9.5  & 1.1  & 8.3  & 1.4  \\
%                                     \cmidrule(l){1-10}
% % ChatGLM-6B (LoRA)     & 17.8 & 3.7  & 10.2 & 25.0 & 3.2  & 10.5 & 29.9 & 13.7 & 24.2 \\
% Baichuan-13B-Chat (LoRA)   & 21.2 & 18.4 & 20.6 & 27.6 & 15.3 & 23.7 & 13.7 & 13.4 & 13.6 \\
                                        \bottomrule
\end{tabular}
\caption{
The automatically evaluated performance of LLMs and all baselines in CGEC. }
\label{Table: Main_Results_CGEC}
\end{table*}

\begin{table*}[ht]
\tiny
\centering
\begin{tabular}{lccccccccc}
\toprule
\multicolumn{1}{c}{\multirow{2}{*}{Model}} & \multicolumn{5}{c}{\textbf{CSC}}                 & \multicolumn{3}{c}{\textbf{CGEC}} & \multicolumn{1}{c}{\textbf{All}} \\
\cmidrule(l){2-6} \cmidrule(l){7-9} \cmidrule(l){10-10}
\multicolumn{1}{c}{}                       & \textbf{SIGHAN15} & \textbf{LAW}  & \textbf{MED}  & \textbf{ODW}  & \textbf{MCSCSet} & \textbf{NLPCC}  & \textbf{MuCGEC} & \textbf{NaCGEC} & \textbf{Avg. Score}              \\
\midrule
gpt-3.5-turbo                              & 18.2     & 34.3 & \ 27.2 & 48.1 & \ 25.0    & 15.9   & 22.0   & 6.3    & \ 24.6                    \\
text-davicin-003                           & \ 19.5     & \ 28.4 & 14.6 & 36.3 & 15.0    & \ 20.2   & 28.6   & \ 7.2    & 21.2                    \\
\midrule
Baichuan-13B-Chat                          & 7.7      & 21.8 & 13.9 & 28.9 & 16.7    & 7.3    & 9.5    & 1.4    & 13.4                    \\
\quad- LoRA + CTC  & 25.6$^{\uparrow 17.9}$& 45.5$^{\uparrow 23.7}$ & 34.8$^{\uparrow 20.9}$ & \ 42.2$^{\uparrow 13.3}$ & 40.5$^{\uparrow 23.8}$    & 20.6$^{\uparrow 13.3}$   & \ 23.7$^{\uparrow 14.2}$   & 13.6$^{\uparrow 12.2}$   & 30.8$^{\uparrow 17.4}$                   \\
\quad- LoRA + CTC w/o few shot  & 26.7$^{\uparrow 19.0}$ & 43.8$^{\uparrow 22.0}$ & 36.4$^{\uparrow 22.5}$ & 39.3$^{\uparrow 10.4}$ & 36.8$^{\uparrow 20.1}$ & 17.9$^{\uparrow 10.6}$ & 19.7$^{\uparrow 10.2}$ & 9.8$^{\uparrow 8.4}$ & 28.8$^{\uparrow 15.4}$                    \\
\quad- LoRA + CTC + General   & 17.9$^{\uparrow 10.2}$ & 28.9$^{\uparrow 7.1}$ & 23.2$^{\uparrow 9.3}$ & 29.7$^{\uparrow 0.8}$ & 18.2$^{\uparrow 1.5}$ & 14.8$^{\uparrow 7.5}$ & 17.3$^{\uparrow 7.8}$ & 5.1$^{\uparrow 3.7}$ & 19.4$^{\uparrow 6.0}$ \\
\quad- FT + CTC  & 39.3$^{\uparrow 31.6}$ & 58.7$^{\uparrow 36.9}$ & 48.2$^{\uparrow 34.3}$ & 47.1$^{\uparrow 18.2}$ & 46.3$^{\uparrow 29.6}$ & 12.2$^{\uparrow 4.9}$  & 10.2 $^{\uparrow 0.7}$ & 31.1$^{\uparrow 29.7}$ &  36.6$^{\uparrow 23.2}$                    \\
\quad- FT + CTC w/o few shot  & 35.2$^{\uparrow 27.5}$ & 57.3$^{\uparrow 35.5}$ & 41.6$^{\uparrow 27.7}$ & 44.9$^{\uparrow 16.0}$ & 40.2$^{\uparrow 23.5}$ & 10.2$^{\uparrow 2.9}$ & 10.0$^{\uparrow 0.5}$ & 32.3$^{\uparrow 30.9}$ & 34.0$^{\uparrow 20.6}$ \\
\quad- FT + CTC + General  & 27.8$^{\uparrow 20.1}$ & 69.8$^{\uparrow 48.0}$ & 52.4$^{\uparrow 38.5}$ & 46.9$^{\uparrow 18.0}$ & 51.3$^{\uparrow 34.6}$ & 17.6$^{\uparrow 10.3}$ & 15.2$^{\uparrow 5.7}$ & 33.8$^{\uparrow 32.4}$ & 39.3$^{\uparrow 25.9}$ \\
\bottomrule
\end{tabular}
\caption{
The performance of fine-tuned LLMs. We use three different datasets including CTC, CTC w/o few shot, and CTC + General. CTC only includes Chinese Text Correction task-specific data, also with few shots examples. General denotes the general data \textit{alpaca\_gpt4\_data\_zh}. FT stands for ``full parameters fine-tuning" }
\label{Table: Main_Results_FineTune}
\end{table*}

\subsection{Main Results}
Table~\ref{Table: Main_Results_CSC}, Table~\ref{Table: Main_Results_CGEC} and Table~\ref{Table: Main_Results_FineTune} show the automatic evaluation results of LLMs and baselines on various datasets. We observe that:
\begin{enumerate}
    \item For CGEC and CSC, despite our careful constraints on the output of LLMs, it still falls far short of the performance of fine-tuned small models on all datasets and all automatic evaluation metrics. This phenomenon shows to a certain extent that for the currently very popular LLMs represented by ChatGPT, the Chinese Text Correction scene is still very challenging and hard.
    \item For the CGEC task, we see that the model performs particularly poorly on NaCGEC, which indicates that the dataset for native Chinese speakers is more challenging than those for foreign language learners, so CGEC for native Chinese speakers will be a research direction worth studying in the future.
    \item For different variants of ChatGPT, we find that in the Chinese Text Correction scenario, \texttt{text-davinci-003} has stronger processing capabilities than \texttt{gpt-3.5-turbo} for daily and general texts (e.g., the SIGHAN15 dataset and CGEC datasets), while for special or data-scarce text domains (e.g., the LAW and MED datasets), \texttt{gpt-3.5-turbo} performed better than \texttt{text-davinci-003}. According to the facts that OpenAI has disclosed to the community, \texttt{gpt-3.5-turbo} has made special optimizations for chat applications compared to \texttt{text-davinci-003}. We guess that this optimization gives ChatGPT a stronger domain adaptation capability.
    \item In the context of fine-tuning LLM, we conduct a comparison between two approaches: LoRA and Full Fine-tuning. We discover that full parameters fine-tuning disrupts a portion of the model's language capabilities. However, it makes it easier for the model to memorize various information present in the fine-tuning data, such as formatting and linguistic forms. This explains why Full fine-tuning achieves high scores in the NaCGEC. NaCGEC, along with our fine-tuning data FCGEC, are collected from native speakers, and they exhibit a high degree of similarity in terms of linguistic characteristics and patterns. When it comes to tasks that require strict adherence to formatting, such as the CSC task, the performance improvement with Full Fine-tuning becomes more pronounced. On the other hand, for tasks that demand a higher level of language proficiency, like the CGEC task, LoRA fine-tuning demonstrates a more significant performance boost.
    \item When it comes to the utilization of fine-tuning data, there are a couple of important considerations. Firstly, the incorporation of samples with few-shot examples in the fine-tuning dataset proves to be beneficial for both LoRA and Full Fine-tuning approaches. Secondly, the inclusion of general data has a more significant positive impact on Full Fine-tuning, while it has a negative effect on LoRA fine-tuning. 
\end{enumerate}

\begin{table*}[ht]
\small
\centering
\begin{tabular}{@{}l|cccc|cccc@{}}
\toprule
\multirow{2}{*}{\textbf{MODEL}}                & \multicolumn{4}{c|}{\textbf{High-Quality}}          & \multicolumn{4}{c}{\textbf{Low-Quality}}          \\
                                    & \textbf{Strict}       & \textbf{Middle1}     & \textbf{Middle2}    & \textbf{Soft}       & \textbf{Strict}       & \textbf{Middle1}    & \textbf{Middle2}     & \textbf{Soft}       \\ 
                             \midrule
REALISE                   & 89.9   & 92.2   & 90.8    & 93.4 & 8.2    & 8.2    & 9.4     & 9.7  \\
text-davinci-003 (0 shot) & 49.5   & 72.0   & 57.1    & 82.8 & 30.5   & 31.4   & 39.3    & 40.2 \\
text-davinci-003 (2 shot) & 56.4   & 73.5   & 61.9    & 80.8 & 27.2   & 27.8   & 34.4    & 35.0 \\
gpt-3.5-turbo (0 shot)    & 54.1   & 78.7   & 60.1    & 87.5 & 32.9   & 34.7   & 43.8    & 45.6 \\
gpt-3.5-turbo (2 shot)    & 68.9   & 78.9   & 73.3    & 84.7 & 28.7   & 30.2   & 34.7    & 36.3 \\
                    \midrule
Baichuan-13B-Chat (LoRA)  & 71.5 & 77.0 & 77.6 & 88.0 & 49.8 & 50.5 & 55.0 & 56.8 \\ 
Baichuan-13B-Chat (FT)  & 77.5 & 82.4 & 81.0 & 90.5 & 38.7 & 39.6 & 40.2 & 42.6 \\
                            \bottomrule
\end{tabular}
\caption{
Human evaluation on the SIGHAN15 dataset. We report the proportion of samples in each category to the total samples in the dataset. Considering the quality of the dataset affects the results a lot, we split the dataset into two parts using the correctness of the source sentence and ground-truth sentence. \textbf{High-Quality} means that ground-truth sentences are fluent and without grammatical errors. \textbf{Low-Quality} indicates that the ground-truth sentence contains grammatical errors. To evaluate the quality of the output sentences, we employ three dimensions: correctness, consistency, and necessity. \textbf{Strict} denotes that the output sentence is without typos, maintains consistency with the source sentence in terms of meaning, and the edits are necessary. \textbf{Middle1} means the output sentence is without typos and maintains consistency with the source sentence, but necessity is not considered. On the other hand, \textbf{Middle2} indicates that the output sentence is without typos and the edits are necessary, but consistency is not considered. \textbf{Soft} means the output sentence is without typos.
}
\label{Table: Human_Results_CSC}
\end{table*}

% Human evaluation CGEC
\begin{table}[ht]
\tiny
% \scriptsize
% \footnotesize
\centering
\begin{tabular}{@{}l|ccc@{}}
\toprule
\textbf{MODEL}                 & \textbf{Correctness}    & \textbf{Consistency}     & \textbf{Crt \& Csit}       \\ 
                             \midrule
Seq2Edit-Baseline                & 40.3   & 85.7   & 35.3      \\    
text-davinci-003 (0 shot) & 41.1 & 74.8 & 36.1  \\
text-davinci-003 (2 shot) & 38.7 & 77.3 & 33.6  \\
gpt-3.5-turbo (0 shot)    & 66.4 & 57.1 & 33.6  \\
gpt-3.5-turbo (2 shot)    & 62.2 & 63.0 & 38.7  \\
                        \midrule
Baichuan-13B-Chat (LoRA)  & 43.2    & 86.4    & 37.3   \\
Baichuan-13B-Chat (FT)   & 9.3   & 97.5   & 9.3   \\
                            \bottomrule
\end{tabular}
\caption{
Human evaluation on the MuCGEC dataset. We report the proportion of samples in each category to the total samples in the dataset. \textbf{Crt \& Csit} implies the attainment of both these qualities simultaneously.
}
\label{Table: Human_Results_CGEC}
\end{table}

\begin{figure*}
\centering
\includegraphics[width=1.8\columnwidth]{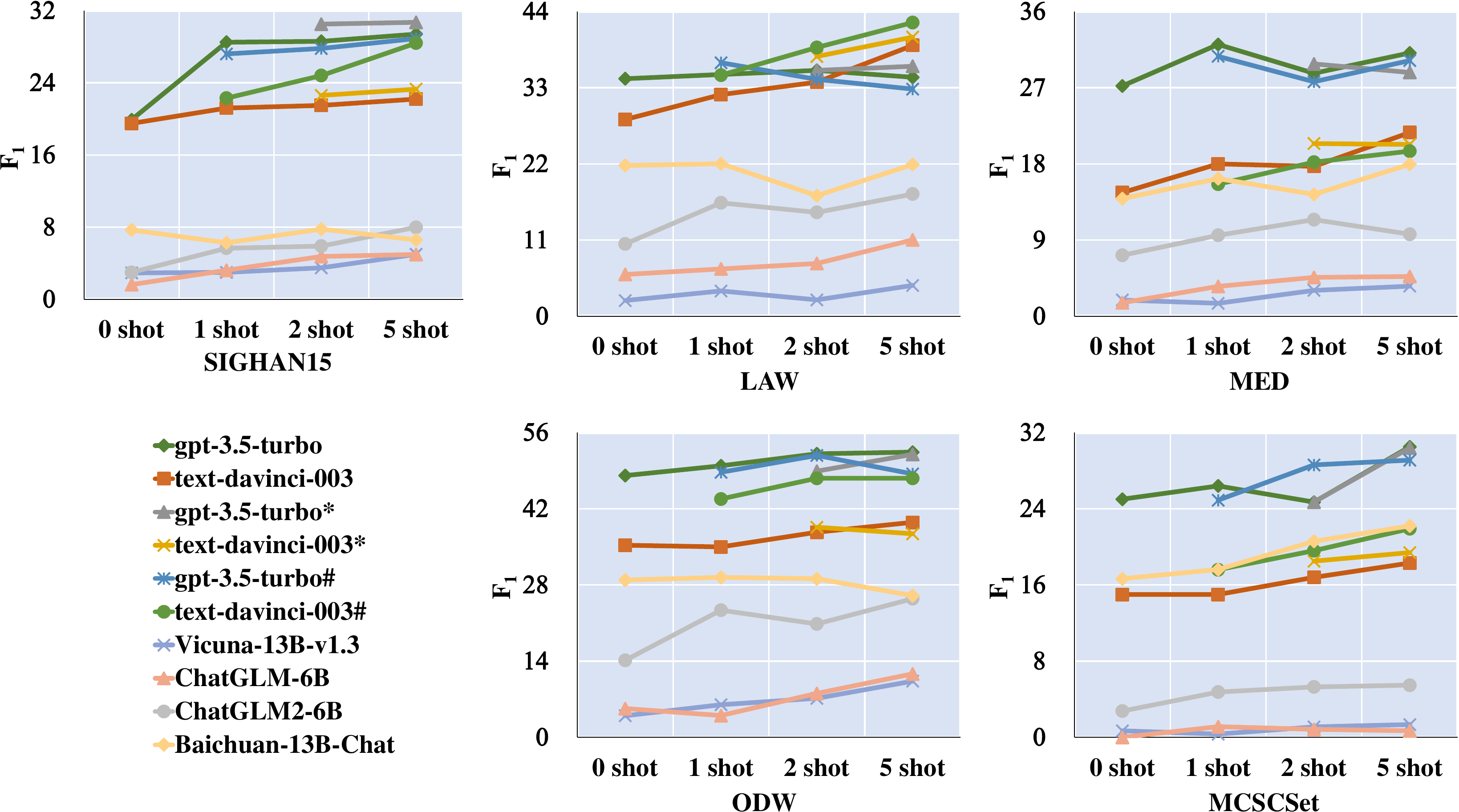}
\caption{ The experiments of in-context learning strategies on the CSC task. We select the correction $F_1$ score to plot the chart. The * means with \textbf{Select correct and erroneous samples} in-context learning strategy. The \# means with \textbf{Select hard erroneous samples} in-context learning strategy.  }
\label{Figure: CSC_Incontext}
\end{figure*}

\subsection{Human Evaluation}
After assessing the performance of LLMs using automated metrics, it is crucial to complement these objective measures with human evaluation. While automated metrics play an important role in model evaluation, they still have a lot of inherent limitations. It is not acceptable for sentence lengths to be modified when calculating automation metrics in the CSC task. Different from non-autoregressive architecture, the decoder-only LM architecture makes the length of the sentence hard to control. For example, if the input sentence is ``今天是重要的天'', the LLMs output will probably change the sentence to ``今天是重要的日子''. The output sentence will be determined as wrong by automated metrics, while we think it is correct based on the commonsense. Therefore, human evaluation is necessary because it can eliminate this kind of bias caused by automated metrics. 

Specifically, for the CSC task, we employ three dimensions to judge each output sentence: correctness, consistency, and necessity. Correctness means the output sentence is fluent and without typos. Consistency means the output sentence can not change the meaning of the input sentence. Necessity means the edits are necessary. For the CGEC task, we employ two dimensions to judge each output sentence: correctness and consistency. Correctness refers to the output sentence without grammatical errors. Consistency indicates that the meaning of the output sentence aligns with the meaning of the references.

Table~\ref{Table: Human_Results_CSC} and Table~\ref{Table: Human_Results_CGEC} present our human evaluation results for CSC and CGEC, respectively. Compared with the results of the automatic evaluation, we know that the human evaluation results show that the error correction ability of LLMs is not so far from that of traditional fine-tuned models. This shows that the traditional automatic evaluation metrics widely used in Chinese Text Correction tasks cannot truly and objectively reflect the correction ability of LLMs, and the design and development of new evaluation metrics for LLMs is an important and valuable future direction.

We also have different observations on the two sub-tasks according to human evaluation results. For the CSC task, the gap between the performance of LLMs on the high-quality samples and the low-quality samples is not as significant as that of the small fine-tuned model, which indicates that LLMs are more fault-tolerant to input data, while traditional fine-tuned small models are more sensitive to the quality of input data. Additionally, for high-quality samples, the models perform better when the necessity is not considered than when the consistency is not considered, while for the low-quality samples, the models perform better when the consistency is not considered than when the necessity is not considered. For high-quality samples, the model performs better when the necessity is not considered than when the continuity is not considered, while for the low-quality samples, the model performs better when the continuity is not considered than when the necessity is not considered. This phenomenon shows that when the sentence is high-quality, the challenge of the CSC task is how to reduce unnecessary editing, and when the sentence quality is poor, the challenge of the CSC task is how to keep the meaning of the source sentence as much as possible. After applying both LoRA and Full Fine-tuning to the model, improvements are observed in settings involving both low-quality and high-quality data. In scenarios with high-quality samples, Full Fine-tuning exhibits a slight advantage over LoRA. However, in settings with low-quality samples, LoRA significantly outperforms Full Fine-tuning.
For the CGEC task, we see that LLMs sometimes perform better than the traditional fine-tuned small model, according to the results of human evaluation. However, the performance of LLMs on the Consistency metric is steadily worse than that of the small model. This also proves our view that when LLMs like ChatGPT perform text error correction, it always tends to play freely, so that sometimes it will change the meaning of the source sentence. Therefore, to make LLMs well adapted to the CGEC task, it is necessary to study how to make LLMs perform more controllable content generation. After the fine-tuning process, the model that used the LoRA technique surpassed the current baseline in the Crt \& Csit tasks. However, in the case of the model that used Full Fine-tuning, its language proficiency appears to have been reduced, making it challenging to generate correct sentences.

\subsection{Analyses and Discussions}

\begin{figure*}
\centering
\includegraphics[width=1.8\columnwidth]{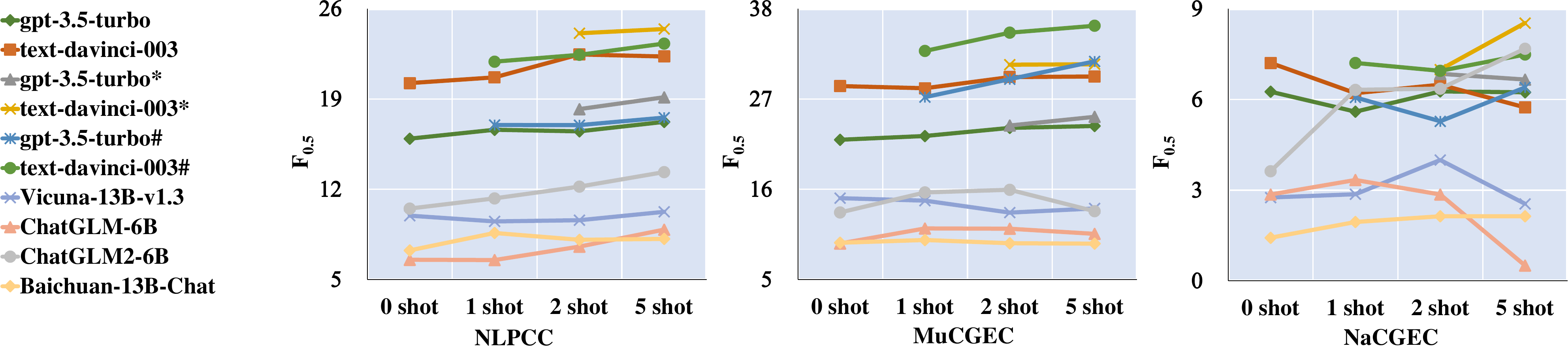}
\caption{The experiments of in-context learning strategies on the CGEC task. We select the $F_{0.5}$ score to plot the chart. The * means with \textbf{Select correct and erroneous samples} in-context learning strategy. The \# means with \textbf{Select hard erroneous samples} in-context learning strategy.  }
\label{Figure: CGEC_Incontext}
\end{figure*}

\noindent\textbf{Effect of Different In-context Learning Strategies}
Figure~\ref{Figure: CSC_Incontext} and Figure~\ref{Figure: CGEC_Incontext} show the automatic evaluation results of LLMs with different in-context learning strategies. In most cases of Figure~\ref{Figure: CSC_Incontext} and Figure~\ref{Figure: CGEC_Incontext}, the model performance is the best when selecting hard erroneous samples, followed by when selecting correct and erroneous samples, and the performance improvement when selecting random erroneous samples is the weakest. Such experimental results reflect the effectiveness of our in-context learning strategies designed for the Chinese Text Correction scenario. Particularly, our experiments show that adding a certain proportion of correct samples to the example samples of the in-context learning of LLMs can bring more performance improvements. We believe that this will provide some guiding significance for the application of LLMs to the Chinese Text Correction scenario in the future. 

\begin{figure}
\centering
\includegraphics[width=0.9\columnwidth]{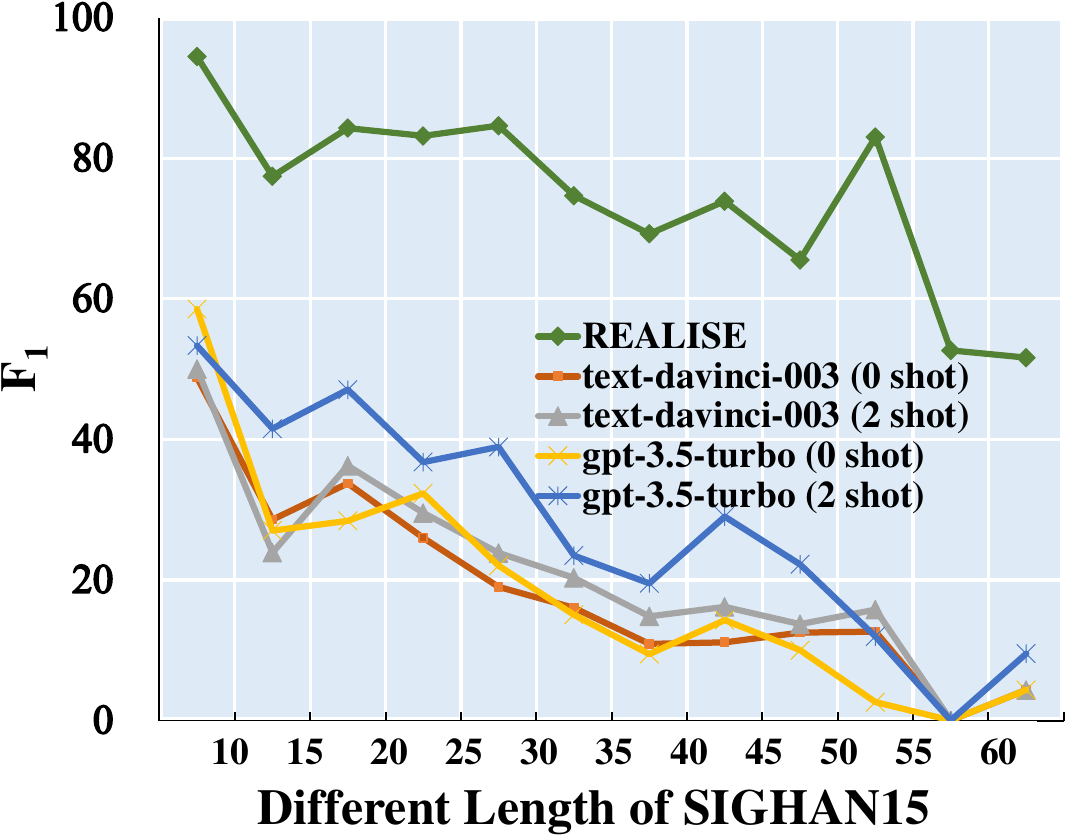}
\caption{The experiments of how the sentence length impacts the model performance on CSC. We select the correction $F_1$ score to plot the chart.  }
\label{Figure: CSC_Length}
\end{figure}

% Different typo's num CSC
\begin{table}[ht]
\scriptsize
\centering
\begin{tabular}{@{}l|cc|cc|cc@{}}
\toprule
\multirow{2}{*}{\textbf{MODEL}}                & \multicolumn{2}{c|}{\textbf{N\_T $=$ 1}}          & \multicolumn{2}{c|}{\textbf{N\_T $=$ 2}}  & \multicolumn{2}{c}{\textbf{N\_T $\geq$ 3}}         \\
                                    & \textbf{D-F}       & \textbf{C-F}                  & \textbf{D-F}       & \textbf{C-F}      & \textbf{D-F}       & \textbf{C-F}               \\ 
                             \midrule

REALISE                         & 91.0         & 90.1        & 66.3         & 62.8        & 42.9                & 39.3               \\
text-davinci-003 (0 shot)         & 41.2         & 32.4        & 30.2         & 11.2        & 10.7                & 3.6                \\
text-davinci-003 (2 shot)         & 47.5         & 35.5        & 27.1         & 12.4        & 10.7                & 3.6                \\
gpt-3.5-turbo (0 shot)            & 37.8         & 31.1        & 26.8         & 19.0        & 7.1                 & 3.6                \\
gpt-3.5-turbo (2 shot)            & 47.5         & 40.9        & 37.2         & 31.4        & 25.0                & 10.7              \\
                                        \bottomrule
\end{tabular}
\caption{
The automatically evaluated performance of LLMs and baseline on SIGHAN15. We divide the dataset into three categories based on the number of typos in a sentence. \textbf{N\_T} stands for the number of typos in a sentence. }
\label{Table: diff typo's num}
\end{table}

\noindent\textbf{Effect of Sample Difficulty}
To further study the Chinese error correction ability of LLMs, we observed the performance of the large model for samples of different difficulties. Specifically, for the CSC task, we measure the difficulty of a sample according to the length of the source sentence and how many typos (i.e., error Chinese characters) it contains. It is worth noting that for the CGEC task, measuring the difficulty of a sample is an overly subjective process~\cite{ye2023cleme}, so we do not involve the CGEC task in this part of the experiment.

Table~\ref{Table: diff typo's num} shows the performance of the model for samples containing different numbers of typos, and Figure~\ref{Figure: CSC_Length} shows how the model performance varies with the length of the input sentence. We see that the performance of all models decreases when the input sentences become more difficult, i.e., contain more typos or are longer. Interestingly, the performance degradation of LLMs is significantly larger than that of the traditional fine-tuned model. We think that maybe LLMs cannot handle multiple errors in a sentence well when dealing with the CSC task, which requires the model to have strong reasoning ability. At the same time, we find that the performance degradation of LLMs with in-context learning is slower than that of LLMs without in-context learning, which suggests that an effective in-context learning strategy can improve the ability of LLMs to handle complex CSC samples to a certain extent.

% Different typo's type CSC
\begin{table*}[ht]
\small
\centering
\begin{tabular}{@{}l|cc|cc|cc|cc@{}}
\toprule
\multirow{2}{*}{\textbf{MODEL}}                & \multicolumn{2}{c|}{\textbf{P\_E}}          & \multicolumn{2}{c|}{\textbf{M\_E}}  & \multicolumn{2}{c|}{\textbf{M\_E\&P\_E}}      & \multicolumn{2}{c}{\textbf{OTHERS}}   \\
                                    & \textbf{D-F}       & \textbf{C-F}                  & \textbf{D-F}       & \textbf{C-F}      & \textbf{D-F}       & \textbf{C-F}        & \textbf{D-F}       & \textbf{C-F}        \\ 
                             \midrule

REALISE                         & 82.0        & 80.5       & 93.3        & 93.3       & 87.5           & 86.8          & 92.3         & 87.2        \\
text-davinci-003 (0 shot)         & 30.5        & 22.0       & 41.4        & 34.5       & 51.8           & 38.0          & 48.8         & 29.3        \\
text-davinci-003 (2 shot)         & 36.3        & 25.2       & 69.0        & 34.5       & 51.2           & 39.7          & 48.8         & 29.3        \\
gpt-3.5-turbo (0 shot)            & 30.4        & 23.9       & 55.2        & 41.4       & 42.6           & 35.5          & 24.4         & 19.5        \\
gpt-3.5-turbo (2 shot)            & 41.9        & 35.6       & 61.5        & 38.5       & 46.9           & 40.1          & 55.0         & 50.0       \\
                                        \bottomrule
\end{tabular}
\caption{
The automatically evaluated performance of LLMs and baseline on SIGHAN15. We divide the dataset into four categories based on the types of typos found in the sentences. The \textbf{P\_E} means the typo is a phonological error, while \textbf{M\_E} means the typo is a morphological error. The \textbf{M\_E\&P\_E} represents sentences with a typo that is both a phonological error and a morphological error. \textbf{OTHERS} indicates that it does not fall into any of the above error categories.}
\label{Table: diff typo's types}
\end{table*}

% Different type's CGEC
\begin{table}[ht]
\tiny
% \scriptsize
\centering
\begin{tabular}{@{}l|ccc|ccc@{}}
\toprule
\multirow{2}{*}{\textbf{MODEL}}                & \multicolumn{3}{c|}{text-davinci-003 (2 shot)}           & \multicolumn{3}{c}{gpt-3.5-turbo (2 shot)}         \\
\cmidrule(l){2-7}
                                    & \textbf{P}       & \textbf{R}                  & \textbf{F}       & \textbf{P}      & \textbf{R}       & \textbf{F}               \\ 
                             \midrule

Structural Confusion & 8.8 & 9.2 & 8.9 & 11.6 & 21.6 & 12.8 \\
Improper Word Order  & 2.2 & 1.9 & 2.1 & 5.9  & 6.7  & 6.1  \\
Missing Component    & 7.0 & 7.5 & 7.1 & 4.3  & 8.8  & 4.8  \\
Improper Collocation & 7.3 & 7.4 & 7.3 & 2.6  & 3.2  & 2.7  \\
Improper Logicality  & 8.8 & 7.7 & 8.6 & 7.1  & 10.7 & 7.6  \\
Redundant Component  & 9.9 & 8.6 & 9.6 & 5.8  & 9.9  & 6.4  \\
                                        \bottomrule
\end{tabular}
\caption{
The automatically evaluated performance of LLMs on different types of grammatical errors on the NaCGEC dataset. }
\label{Table: diff categories CGEC}
\end{table}

\noindent\textbf{Fine-grained Performance Analysis}
To deeply explore the real performance of LLMs for the CSC and CGEC tasks, We report in detail the performance of LLMs when handling fine-grained error types in Table~\ref{Table: diff typo's types} and Table~\ref{Table: diff categories CGEC}. Judging from the fine-grained performance of CSC in Table~\ref{Table: diff typo's types}, LLMs handle morphological errors better than phonological errors. This experimental result can be explained from the perspective of linguistics. Because Chinese characters are pictographic characters, the meaning of most Chinese characters is closely related to their shape and stroke structure, but the connection with their pinyin pronunciation is relatively weak. Therefore, while LLMs such as ChatGPT can capture the meaning of Chinese characters, they also capture the structural characteristics of Chinese characters in a sense, which leads to their better performance in processing morphological errors. Of course, this explanation is just a reasonable guess based on the experimental results. After all, ChatGPT is still a black box, and its interpretability research in the Chinese Text Correction field is also a very interesting direction in the future. 
As for the CGEC results in Table~\ref{Table: diff categories CGEC}, we can see that \texttt{text-davinci-003} has the worst performance for the type of ``Improper Word Order'', and \texttt{gpt-3.5-turbo} has the worst performance for the type of ``Improper Collocation''. This reflects how improving the generalization performance of LLMs among various fine-grained grammatical errors will be very challenging.

\noindent\textbf{Reasons for the Errors of LLMs: }
We analyze the results of LLMs and conclude some main error reasons: 
\begin{itemize}
    \item The typos in the sentence change the semantics of the sentence badly. It is even difficult for humans to correct this kind of typos. For example, ``\textcolor{red}{遭}(\textcolor{blue}{超})级市场''. The typo in the sentence makes it hard to fix.
    \item Lacking fine-grained knowledge such as some specific location names appearing in the dataset. For example, ``\textcolor{red}{市}大'' and ``\textcolor{blue}{师}大''. It can not be corrected without specific training data or fine-grained knowledge.
    \item The training data distribution of LLMs will influence their expression habits. They will correct the uncommon but correct expression to satisfy its habits. For example, ``交\textcolor{blue}{往}多少朋友'' will be corrected as ``交\textcolor{blue}{谈}多少朋友'', which is called overcorrection.
    \item Lack of multimodal information. In the CSC task, morphological and phonological information about characters is crucial. We find that LLMs know the pinyin of Chinese characters, but they can not understand how to pronounce it, so it is hard for LLMs to correct phonological errors. For example, ChatGPT prefers to correct ``参加\textcolor{red}{误}会'' as ``参加\textcolor{red}{宴}会'' instead of ``参加\textcolor{blue}{舞}会''.
    \item LLMs tend to generate fluent output, but often alter the original meaning of input sentences incorrectly. For instance, ``孩子的兴趣\textcolor{red}{无一没有}跟父母的兴趣\textcolor{blue}{相干}(All of the child's interests are related to the parents)'' is corrected as ``孩子的兴趣\textcolor{red}{没有一个}与父母的兴趣\textcolor{red}{相关}(None of the child's interests are related to the parents)'', instead of ``孩子的兴趣\textcolor{blue}{无一不是}跟父母的兴趣\textcolor{blue}{相干}''.
    \item In certain cases, LLMs encounter challenges in striking a balance between the precision and the imperative of minimizing modifications in text error correction, thereby resulting in under-correction (or incomplete correction). For example, ``我\textcolor{red}{从个个的经验}来谈这个题目'' should be corrected to ``我\textcolor{blue}{从个人的角度}来谈这个题目(I will discuss this topic from my perspective)'' or ``我\textcolor{blue}{根据个人的经验}来谈这个题目(I will discuss this topic based on my experience)'', but LLMs erroneously revise it to ``我\textcolor{blue}{从个人的经验}来谈这个题目''.
\end{itemize}

% Case Study CSC
\begin{table}[ht]
\tiny
% \scriptsize
% \footnotesize
\centering
% \scalebox{0.9}{
\begin{tabular}{@{}c|l@{}}
\toprule
\multicolumn{1}{l|}{}   & Input: \textcolor{red}{情}给李小姐打电话。 \\
\multicolumn{1}{l|}{}        & Translation: \textcolor{blue}{Feeling} call Miss Li.        \\
\multicolumn{1}{c|}{\textbf{Missing}}        & Output: \textcolor{red}{情}给李小姐打电话。         \\
\multicolumn{1}{c|}{\textbf{Detection}}        & Translation: \textcolor{blue}{Feeling} call Miss Li.        \\
\multicolumn{1}{l|}{}        & Correct:\textcolor{blue}{请}给李小姐打电话。         \\
\multicolumn{1}{l|}{}        & Translation: \textcolor{blue}{Please} call Miss Li.        \\
\midrule
\multicolumn{1}{l|}{}   & Input: 他\textcolor{red}{收}到山上的时候，非常高兴。\\
\multicolumn{1}{l|}{}        & Translation:  When he \textcolor{red}{received} up the mountain, he was very happy.\\
\multicolumn{1}{c|}{\textbf{Wrong}}        & Output: 他\textcolor{red}{收}到\textcolor{red}{信件}的时候，非常高兴。         \\
\multicolumn{1}{c|}{\textbf{Direction}}        & Translation:  When he \textcolor{red}{received} the \textcolor{red}{mail}, he was very happy. \\
\multicolumn{1}{l|}{}        & Correct: 他\textcolor{blue}{走}到山上的时候，非常高兴。         \\
\multicolumn{1}{l|}{}        & Translation: When he \textcolor{blue}{walked} up the mountain, he was very happy.        \\
\midrule
\multicolumn{1}{l|}{}   & Input: 他们的\textcolor{red}{吵翻}很不错，再说他们做的咖哩鸡也好吃！ \\
\multicolumn{1}{l|}{}        & Translation: Their \textcolor{red}{quarrel} is pretty good, and their chicken curry is  \\
\multicolumn{1}{l|}{}        & \qquad\qquad\quad delicious too!       \\
\multicolumn{1}{c|}{\textbf{Inconsistent}}        & Output: 他们的\textcolor{red}{巧舌如簧}很不错，再说他们做的咖哩鸡也好吃！         \\
\multicolumn{1}{c|}{\textbf{Context}}        & Translation: Their \textcolor{red}{glib} is pretty good, and their chicken curry is  \\
\multicolumn{1}{l|}{}        & \qquad\qquad\quad delicious too!       \\
\multicolumn{1}{l|}{}        & Correct:他们的\textcolor{blue}{炒饭}很不错，再说他们做的咖哩鸡也好吃！         \\
\multicolumn{1}{l|}{}        & Translation: Their \textcolor{blue}{fried rice} is pretty good, and their chicken curry is  \\
\multicolumn{1}{l|}{}        & \qquad\qquad\quad delicious too!       \\
\midrule
\multicolumn{1}{l|}{}   & Input: 这次\textcolor{red}{署}假小花去台北旅行顺便去看她的男朋友。 \\
\multicolumn{1}{l|}{}        & Translation: This \textcolor{red}{sign} holiday, Xiaohua went to Taipei to visit \\
\multicolumn{1}{l|}{}        & \qquad\qquad\quad her boyfriend by the way.       \\
\multicolumn{1}{c|}{\textbf{Incorrect}}        & Output: 这次\textcolor{red}{叔}假小花去台北旅行顺便去看她的男朋友。         \\
\multicolumn{1}{c|}{\textbf{Expression}}        & Translation: This \textcolor{red}{uncle} holiday, Xiaohua went to Taipei to visit \\
\multicolumn{1}{l|}{}        & \qquad\qquad\quad her boyfriend by the way.       \\
\multicolumn{1}{l|}{}        & Correct: 这次\textcolor{blue}{暑}假小花去台北旅行顺便去看她的男朋友。         \\
\multicolumn{1}{l|}{}        & Translation: This \textcolor{blue}{summer} holiday, Xiaohua went to Taipei to visit \\
\multicolumn{1}{l|}{}        & \qquad\qquad\quad her boyfriend by the way.       \\
\bottomrule
\end{tabular}
% }
\caption{ 
Cases from the SIGHAN15 test set. The ``Output'' is from \texttt{gpt-3.5-turbo} 2 shot. We categorized the types of model errors into the following categories: \textbf{Missing Detection}, \textbf{Wrong Direction}, \textbf{Inconsistent Context}, and \textbf{Incorrect Expression}. We mark the \textcolor{red}{wrong}/\textcolor{blue}{correct} characters.  }
\label{Table: Case_Studies_CSC}
\end{table}

% Case Study CGEC
\begin{table*}[ht]
\tiny
\centering
\begin{tabular}{@{}c|l@{}}
\toprule
\multicolumn{1}{l|}{}   & Input: 我们改变了我们的地球，所以我们要\textcolor{red}{保护}这个地球的\textcolor{red}{自然}。 \\
\multicolumn{1}{l|}{}        & Translation: We changed our earth, so we need to \textcolor{red}{protect} the earth's \textcolor{red}{natural}.        \\
\multicolumn{1}{c|}{\textbf{Improper}}        & Output: 我们改变了地球，所以我们要\textcolor{red}{保护}地球的\textcolor{red}{自然}。         \\
\multicolumn{1}{c|}{\textbf{Collocation}}        & Translation: We changed the earth, so we need to \textcolor{red}{protect} the earth's \textcolor{red}{natural}.        \\
\multicolumn{1}{l|}{}        & Correct:我们改变了我们的地球，所以我们要\textcolor{blue}{保护}地球的自然\textcolor{blue}{环境}。        \\
\multicolumn{1}{l|}{}        & Translation: We changed our earth, so we need to \textcolor{blue}{protect} the earth's \textcolor{blue}{natural environment}.       \\
\midrule
\multicolumn{1}{l|}{}   & Input: \textcolor{red}{比如说是}生活方式，吃的东西都会有区别。\\
\multicolumn{1}{l|}{}        & Translation:  \textcolor{blue}{For example is}, lifestyle and what you eat will be different.   \\
\multicolumn{1}{c|}{\textbf{Over}}        & Output: \textcolor{blue}{比如说}生活方式和\textcolor{red}{饮食习惯}都会有区别。         \\
\multicolumn{1}{c|}{\textbf{Correction}}        & Translation: \textcolor{blue}{For example}, lifestyle and \textcolor{red}{eating habits} will be different. \\
\multicolumn{1}{l|}{}        & Correct: \textcolor{blue}{比如说}生活方式，吃的东西都会有区别。         \\
\multicolumn{1}{l|}{}        & Translation: \textcolor{blue}{For example}, lifestyle and what you eat will be different.        \\
\midrule
\multicolumn{1}{l|}{}   & Input: 还有公众利益方面\textcolor{red}{来说}，社会上还有不吸烟的人。最近他们的声音越来越大，因为他们虽然不吸烟可是受到吸烟人的影响，\textcolor{red}{他们叫}被吸烟者。\\
\multicolumn{1}{l|}{}        & Translation: Also, \textcolor{red}{in terms} public interest, there are still non-smokers in society. Recently their voices have become louder because although they do not smoke \\
\multicolumn{1}{l|}{}       &\qquad\qquad\qquad but are influenced by smokers, \textcolor{red}{they called} passive smokers.  \\
\multicolumn{1}{c|}{\textbf{Missing}}        & Output: 还有公众利益方面\textcolor{red}{来说}，社会上还有不吸烟的人。最近他们的声音越来越大，因为他们虽然不吸烟，但受到吸烟人的影响，\textcolor{blue}{他们被称为}被动吸烟者。         \\
\multicolumn{1}{c|}{\textbf{Component}}  & Translation: Also, \textcolor{red}{in terms} public interest, there are still non-smokers in society. Recently their voices have become louder because although they do not smoke \\
\multicolumn{1}{l|}{}       &\qquad\qquad\qquad but are influenced by smokers, \textcolor{blue}{they are called} passive smokers. \\
\multicolumn{1}{l|}{}        & Correct: 还有\textcolor{blue}{从}公众利益方面\textcolor{blue}{来说}，社会上还有不吸烟的人。最近他们的声音越来越大，因为他们虽然不吸烟可是还是受到了吸烟人的影响，\textcolor{blue}{他们叫做}被吸烟者。         \\
\multicolumn{1}{l|}{}        & Translation:     Also, \textcolor{blue}{in terms of} public interest, there are still non-smokers in society. Recently their voices have become louder because although they do not smoke\\
\multicolumn{1}{l|}{}       &\qquad\qquad\qquad but are influenced by smokers, \textcolor{blue}{they are called} passive smokers.  \\
\midrule
\multicolumn{1}{l|}{}   & Input: 在房间时\textcolor{red}{砂}能打伞，不然个子长不高。 \\
\multicolumn{1}{l|}{}        & Translation: \textcolor{red}{Sand can} open an umbrella in the room, or you will not be tall. \\
\multicolumn{1}{c|}{\textbf{Improper}}        & Output: 在房间里\textcolor{red}{沙子}能打伞，不然个子长不高。         \\
\multicolumn{1}{c|}{\textbf{Logicality}}  & Translation: \textcolor{red}{Sand can} open an umbrella in the room, or you will not be tall. \\
\multicolumn{1}{l|}{}        & Correct: 在房间时\textcolor{blue}{不}能打伞，不然个子长不高。         \\
\multicolumn{1}{l|}{}        & Translation: \textcolor{blue}{You cannot} open an umbrella in the room, or you will not be tall. \\
\midrule
\multicolumn{1}{l|}{}   & Input: 我也想\textcolor{red}{强烈的}父母或者花心的爱人\textcolor{red}{不是}好的家庭。 \\
\multicolumn{1}{l|}{}        & Translation: I don't think \textcolor{red}{intense} parents or philandering lovers \textcolor{red}{are} a good family. \\
\multicolumn{1}{c|}{\textbf{Under}}        & Output: 我认为\textcolor{red}{强烈的}父母或者花心的爱人\textcolor{blue}{并不构成}一个好的家庭。         \\
\multicolumn{1}{c|}{\textbf{Correction}}  & Translation: I don't think \textcolor{red}{intense} parents or philandering lovers \textcolor{blue}{make} a good family. \\
\multicolumn{1}{l|}{}        & Correct: 我也认为\textcolor{blue}{有性格强势的}父母或者花心的爱人\textcolor{blue}{不能组成}好的家庭。      \\
\multicolumn{1}{l|}{}        & Translation: I don't think \textcolor{blue}{aggressive} parents or philandering lovers \textcolor{blue}{make} a good family. \\
\bottomrule
\end{tabular}
\caption{ 
Cases from the MuCGEC test set. The ``Output'' is from \texttt{gpt-3.5-turbo} 2 shot. We categorized the types of model errors into the following categories: \textbf{Improper Collocation}, \textbf{Over Correction}, \textbf{Missing Component}, \textbf{Improper Logicality} and \textbf{Under Correction}. We mark the \textcolor{red}{wrong}/\textcolor{blue}{correct} characters.  }
\label{Table: Case_Studies_CGEC}
\end{table*}

\subsection{Case Study}
We conduct error case studies for LLMs, as presented in Table~\ref{Table: Case_Studies_CSC} and Table~\ref{Table: Case_Studies_CGEC}. 
\begin{itemize}
    \item \textbf{Missing Detection (CSC)}, refers to instances where the LLMs fail to identify an error in the given sentence.
    \item \textbf{Wrong Direction (CSC)}, pertains to situations where the model corrects the wrong section of the input, resulting in a grammatically correct sentence but altering the intended meaning of the original input.
    \item \textbf{Inconsistent Context (CSC)}, describes cases where models overlook the contextual information when correcting.
    \item \textbf{Incorrect Expression (CSC)}, denotes scenarios where the output sentence still contains typographical or grammatical errors.
    \item \textbf{Improper Collocation (CGEC)}, refers to instances where models produce output that includes incorrect or unnatural combinations of words or phrases, resulting in poor collocation. 
    \item \textbf{Over Correction (CGEC)}, occurs when the model excessively modifies the input sentence, resulting in an over-correction that may introduce errors or change the intended meaning.
    \item \textbf{Missing Component (CGEC)}, refers to situations where the model fails to add necessary components or elements to the sentence, leading to an ambiguous output.
    \item \textbf{Improper Logicality (CGEC)}, refers to instances where the model fails to correctly understand the logical relationship or common sense knowledge in a sentence, resulting in modified sentences that still contain logical errors.
    \item \textbf{Under Correction (CGEC)}, denotes cases where the model inadequately corrects the input, leaving behind errors or failing to address the grammatical or typographical issues present.
\end{itemize}
% The first category, \textbf{Missing Detection}, refers to instances where the LLM fails to identify an error in the given sentence. The second category, \textbf{Wrong Direction}, pertains to situations where the model corrects the wrong section of the input, resulting in a grammatically correct sentence but altering the intended meaning of the original input. The third category, \textbf{Inconsistent Context}, describes cases where the model overlooks the contextual information when making corrections. Lastly, the fourth category, \textbf{Incorrect Expression}, denotes scenarios where the output sentence still contains typographical or grammatical errors.

\section{Related Work}
Chinese Text Correction is a Chinese application closely related to daily life. Due to the complex characteristics of the Chinese language, Chinese Text Correction is a fundamental yet challenging task~\cite{zhao2022overview}. According to different types of errors, Chinese Text Correction is mainly divided into two categories, namely CGEC and CSC, and CGEC is further divided into CSL CGEC and native CGEC based on different target user groups. For the CSL CGEC task, with the Lang8~\cite{10.1007/978-3-319-99501-4_41} and HSK~\cite{zhang2009features} corpora as resources, the NLPCC2018~\cite{10.1007/978-3-319-99501-4_41} and CGED~\cite{rao-etal-2018-overview,rao-etal-2020-overview} are widely used evaluation benchmarks. Besides, considering the subjectivity of the grammatical error correction task and the diversity of correction methods, multi-reference CGEC evaluation datasets such as YACLC~\cite{wang2021yaclc} and MuCGEC~\cite{zhang-etal-2022-mucgec} are also constructed from the Lang8 corpus. However, the above-mentioned datasets are all derived from the grammatical errors made by foreign Chinese learners in their process of learning Chinese. There is a large gap between the language usage habits of foreigners and Chinese native speakers. This gap will cause the models trained or evaluated on these datasets to be unable to adapt well to a wider range of Chinese application scenarios. Therefore, researchers recently begin to pay attention to the CGEC task focusing on Chinese native speakers. The latest native CGEC datasets include FCGEC~\cite{xu-etal-2022-fcgec} and NaCGEC~\cite{ma-etal-2022-linguistic}. Compared with the relatively simple CSL CGEC task, the types of Chinese grammatical errors that the native CGEC task focuses on are more complex, such as structural confusion, improper logicality, missing component, redundant component, improper collection, and improper word order. The CSC task mainly focuses on Chinese spelling errors caused by confusion of pronunciation and strokes. For quite a long time in the past, the SIGHAN datasets~\cite{wu-etal-2013-sighan2013,yu-etal-2014-sighan2014,tseng-etal-2015-sighan2015} are the only evaluation benchmark for the CSC task and are widely used by researchers. Recently, considering the singleness of the SIGHAN datasets, some new CSC datasets have been proposed to evaluate the performance of the CSC model in different domains~\cite{Lv_2023,10.1145/3511808.3557636}.

In the field of Grammatical Error Correction, in addition to Chinese, English Grammar Error Correction (EGEC) is also widely concerned. There have been several works~\cite{wu2023chatgpt,fang2023chatgpt,coyne2023analyzing} to evaluate ChatGPT on the EGEC task. Because the linguistic characteristics of Chinese and English are essentially different, unlike previous work, our work focuses on Chinese and aims to explore the Chinese correction capabilities of LLMs and promote the development and progress of Chinese Text Correction in the era of LLMs.

\section{Conclusion}
In this paper, we analyze the correction ability of the existing LLMs, represented by ChatGPT. We find that the text correction ability of the LLMs still has some gaps with the previous state-of-the-art fine-tuned models. Through human evaluation, we discover that the LLMs demonstrate greater resilience in addressing issues of fluency in the text. Additionally, we observe that as the difficulty increases, the performance of LLMs tends to decline more significantly compared to the fine-tuned small models. It is evident that the Chinese Text Correction capability of LLMs has not yet been fully adapted to the current production environment, necessitating increased attention and resources in this area.

\section*{Acknowledgments}
This research is supported by National Natural Science Foundation of China (Grant No.62276154), Research  Center for Computer Network (Shenzhen) Ministry of Education，the Natural Science Foundation of Guangdong Province (Grant No.2023A1515012914), the Basic Research Fund of Shenzhen City (Grant No. JCYJ20210324120012033 and JSGG20210802154402007), the Shenzhen Science and Technology Program (No. WDZC20231128091437002), the Major Key Project of PCL for Experiments and Applications (PCL2021A06), and Overseas Cooperation Research Fund of Tsinghua Shenzhen International Graduate School (HW2021008).

% {\appendix[Proof of the Zonklar Equations]
% Use $\backslash${\tt{appendix}} if you have a single appendix:
% Do not use $\backslash${\tt{section}} anymore after $\backslash${\tt{appendix}}, only $\backslash${\tt{section*}}.
% If you have multiple appendixes use $\backslash${\tt{appendices}} then use $\backslash${\tt{section}} to start each appendix.
% You must declare a $\backslash${\tt{section}} before using any $\backslash${\tt{subsection}} or using $\backslash${\tt{label}} ($\backslash${\tt{appendices}} by itself
%  starts a section numbered zero.)}


\bibliographystyle{IEEEtran}
\bibliography{IEEEabrv}

\begin{thebibliography}{10}
\providecommand{\url}[1]{#1}
\csname url@samestyle\endcsname
\providecommand{\newblock}{\relax}
\providecommand{\bibinfo}[2]{#2}
\providecommand{\BIBentrySTDinterwordspacing}{\spaceskip=0pt\relax}
\providecommand{\BIBentryALTinterwordstretchfactor}{4}
\providecommand{\BIBentryALTinterwordspacing}{\spaceskip=\fontdimen2\font plus
\BIBentryALTinterwordstretchfactor\fontdimen3\font minus \fontdimen4\font\relax}
\providecommand{\BIBforeignlanguage}[2]{{%
\expandafter\ifx\csname l@#1\endcsname\relax
\typeout{** WARNING: IEEEtran.bst: No hyphenation pattern has been}%
\typeout{** loaded for the language `#1'. Using the pattern for}%
\typeout{** the default language instead.}%
\else
\language=\csname l@#1\endcsname
\fi
#2}}
\providecommand{\BIBdecl}{\relax}
\BIBdecl

\bibitem{zhao2023survey}
W.~X. Zhao, K.~Zhou, J.~Li, T.~Tang, X.~Wang, Y.~Hou, Y.~Min, B.~Zhang, J.~Zhang, Z.~Dong, Y.~Du, C.~Yang, Y.~Chen, Z.~Chen, J.~Jiang, R.~Ren, Y.~Li, X.~Tang, Z.~Liu, P.~Liu, J.-Y. Nie, and J.-R. Wen, ``A survey of large language models,'' 2023.

\bibitem{wei2022emergent}
\BIBentryALTinterwordspacing
J.~Wei, Y.~Tay, R.~Bommasani, C.~Raffel, B.~Zoph, S.~Borgeaud, D.~Yogatama, M.~Bosma, D.~Zhou, D.~Metzler, E.~H. Chi, T.~Hashimoto, O.~Vinyals, P.~Liang, J.~Dean, and W.~Fedus, ``Emergent abilities of large language models,'' \emph{Transactions on Machine Learning Research}, 2022, survey Certification. [Online]. Available: \url{https://openreview.net/forum?id=yzkSU5zdwD}
\BIBentrySTDinterwordspacing

\bibitem{NEURIPS2022_9d560961}
\BIBentryALTinterwordspacing
J.~Wei, X.~Wang, D.~Schuurmans, M.~Bosma, b.~ichter, F.~Xia, E.~Chi, Q.~V. Le, and D.~Zhou, ``Chain-of-thought prompting elicits reasoning in large language models,'' in \emph{Advances in Neural Information Processing Systems}, S.~Koyejo, S.~Mohamed, A.~Agarwal, D.~Belgrave, K.~Cho, and A.~Oh, Eds., vol.~35.\hskip 1em plus 0.5em minus 0.4em\relax Curran Associates, Inc., 2022, pp. 24\,824--24\,837. [Online]. Available: \url{https://proceedings.neurips.cc/paper_files/paper/2022/file/9d5609613524ecf4f15af0f7b31abca4-Paper-Conference.pdf}
\BIBentrySTDinterwordspacing

\bibitem{he2023chatgpt}
M.~He and P.~N. Garner, ``Can chatgpt detect intent? evaluating large language models for spoken language understanding,'' 2023.

\bibitem{wei2023zeroshot}
X.~Wei, X.~Cui, N.~Cheng, X.~Wang, X.~Zhang, S.~Huang, P.~Xie, J.~Xu, Y.~Chen, M.~Zhang, Y.~Jiang, and W.~Han, ``Zero-shot information extraction via chatting with chatgpt,'' 2023.

\bibitem{yang2023exploring}
X.~Yang, Y.~Li, X.~Zhang, H.~Chen, and W.~Cheng, ``Exploring the limits of chatgpt for query or aspect-based text summarization,'' 2023.

\bibitem{liu-etal-2010-visually}
\BIBentryALTinterwordspacing
C.-L. Liu, M.-H. Lai, Y.-H. Chuang, and C.-Y. Lee, ``Visually and phonologically similar characters in incorrect simplified {C}hinese words,'' in \emph{Coling 2010: Posters}.\hskip 1em plus 0.5em minus 0.4em\relax Beijing, China: Coling 2010 Organizing Committee, Aug. 2010, pp. 739--747. [Online]. Available: \url{https://aclanthology.org/C10-2085}
\BIBentrySTDinterwordspacing

\bibitem{zhao2022overview}
H.~Zhao, B.~Wang, D.~Wu, W.~Che, Z.~Chen, and S.~Wang, ``Overview of ctc 2021: Chinese text correction for native speakers,'' 2022.

\bibitem{wang2020comprehensive}
Y.~Wang, Y.~Wang, J.~Liu, and Z.~Liu, ``A comprehensive survey of grammar error correction,'' 2020.

\bibitem{ye2022focus}
J.~Ye, Y.~Li, S.~Ma, R.~Xie, W.~Wu, and H.-T. Zheng, ``Focus is all you need for chinese grammatical error correction,'' \emph{arXiv preprint arXiv:2210.12692}, 2022.

\bibitem{wu-etal-2013-integrating}
\BIBentryALTinterwordspacing
J.-c. Wu, H.-w. Chiu, and J.~S. Chang, ``Integrating dictionary and web n-grams for {C}hinese spell checking,'' in \emph{International Journal of Computational Linguistics {\&} {C}hinese Language Processing, Volume 18, Number 4, {D}ecember 2013-Special Issue on Selected Papers from {ROCLING} {XXV}}, Dec. 2013. [Online]. Available: \url{https://aclanthology.org/O13-5002}
\BIBentrySTDinterwordspacing

\bibitem{li-etal-2022-past}
\BIBentryALTinterwordspacing
Y.~Li, Q.~Zhou, Y.~Li, Z.~Li, R.~Liu, R.~Sun, Z.~Wang, C.~Li, Y.~Cao, and H.-T. Zheng, ``The past mistake is the future wisdom: Error-driven contrastive probability optimization for {C}hinese spell checking,'' in \emph{Findings of the Association for Computational Linguistics: ACL 2022}.\hskip 1em plus 0.5em minus 0.4em\relax Dublin, Ireland: Association for Computational Linguistics, May 2022, pp. 3202--3213. [Online]. Available: \url{https://aclanthology.org/2022.findings-acl.252}
\BIBentrySTDinterwordspacing

\bibitem{zhang2023contextual}
D.~Zhang, Y.~Li, Q.~Zhou, S.~Ma, Y.~Li, Y.~Cao, and H.-T. Zheng, ``Contextual similarity is more valuable than character similarity: An empirical study for chinese spell checking,'' in \emph{ICASSP 2023-2023 IEEE International Conference on Acoustics, Speech and Signal Processing (ICASSP)}.\hskip 1em plus 0.5em minus 0.4em\relax IEEE, 2023, pp. 1--5.

\bibitem{nagata-sakaguchi-2016-phrase}
\BIBentryALTinterwordspacing
R.~Nagata and K.~Sakaguchi, ``Phrase structure annotation and parsing for learner {E}nglish,'' in \emph{Proceedings of the 54th Annual Meeting of the Association for Computational Linguistics (Volume 1: Long Papers)}.\hskip 1em plus 0.5em minus 0.4em\relax Berlin, Germany: Association for Computational Linguistics, Aug. 2016, pp. 1837--1847. [Online]. Available: \url{https://aclanthology.org/P16-1173}
\BIBentrySTDinterwordspacing

\bibitem{xie2022an}
\BIBentryALTinterwordspacing
S.~M. Xie, A.~Raghunathan, P.~Liang, and T.~Ma, ``An explanation of in-context learning as implicit bayesian inference,'' in \emph{International Conference on Learning Representations}, 2022. [Online]. Available: \url{https://openreview.net/forum?id=RdJVFCHjUMI}
\BIBentrySTDinterwordspacing

\bibitem{bansal-etal-2023-rethinking}
\BIBentryALTinterwordspacing
H.~Bansal, K.~Gopalakrishnan, S.~Dingliwal, S.~Bodapati, K.~Kirchhoff, and D.~Roth, ``Rethinking the role of scale for in-context learning: An interpretability-based case study at 66 billion scale,'' in \emph{Proceedings of the 61st Annual Meeting of the Association for Computational Linguistics (Volume 1: Long Papers)}.\hskip 1em plus 0.5em minus 0.4em\relax Toronto, Canada: Association for Computational Linguistics, Jul. 2023, pp. 11\,833--11\,856. [Online]. Available: \url{https://aclanthology.org/2023.acl-long.660}
\BIBentrySTDinterwordspacing

\bibitem{dai-etal-2023-gpt}
\BIBentryALTinterwordspacing
D.~Dai, Y.~Sun, L.~Dong, Y.~Hao, S.~Ma, Z.~Sui, and F.~Wei, ``Why can {GPT} learn in-context? language models secretly perform gradient descent as meta-optimizers,'' in \emph{Findings of the Association for Computational Linguistics: ACL 2023}.\hskip 1em plus 0.5em minus 0.4em\relax Toronto, Canada: Association for Computational Linguistics, Jul. 2023, pp. 4005--4019. [Online]. Available: \url{https://aclanthology.org/2023.findings-acl.247}
\BIBentrySTDinterwordspacing

\bibitem{dong2023survey}
Q.~Dong, L.~Li, D.~Dai, C.~Zheng, Z.~Wu, B.~Chang, X.~Sun, J.~Xu, L.~Li, and Z.~Sui, ``A survey on in-context learning,'' 2023.

\bibitem{wang-etal-2018-hybrid}
\BIBentryALTinterwordspacing
D.~Wang, Y.~Song, J.~Li, J.~Han, and H.~Zhang, ``A hybrid approach to automatic corpus generation for {C}hinese spelling check,'' in \emph{Proceedings of the 2018 Conference on Empirical Methods in Natural Language Processing}.\hskip 1em plus 0.5em minus 0.4em\relax Brussels, Belgium: Association for Computational Linguistics, Oct.-Nov. 2018, pp. 2517--2527. [Online]. Available: \url{https://aclanthology.org/D18-1273}
\BIBentrySTDinterwordspacing

\bibitem{wu-etal-2013-sighan2013}
\BIBentryALTinterwordspacing
S.-H. Wu, C.-L. Liu, and L.-H. Lee, ``{C}hinese spelling check evaluation at {SIGHAN} bake-off 2013,'' in \emph{Proceedings of the Seventh {SIGHAN} Workshop on {C}hinese Language Processing}.\hskip 1em plus 0.5em minus 0.4em\relax Nagoya, Japan: Asian Federation of Natural Language Processing, Oct. 2013, pp. 35--42. [Online]. Available: \url{https://aclanthology.org/W13-4406}
\BIBentrySTDinterwordspacing

\bibitem{yu-etal-2014-sighan2014}
\BIBentryALTinterwordspacing
L.-C. Yu, L.-H. Lee, Y.-H. Tseng, and H.-H. Chen, ``Overview of {SIGHAN} 2014 bake-off for {C}hinese spelling check,'' in \emph{Proceedings of The Third {CIPS}-{SIGHAN} Joint Conference on {C}hinese Language Processing}.\hskip 1em plus 0.5em minus 0.4em\relax Wuhan, China: Association for Computational Linguistics, Oct. 2014, pp. 126--132. [Online]. Available: \url{https://aclanthology.org/W14-6820}
\BIBentrySTDinterwordspacing

\bibitem{tseng-etal-2015-sighan2015}
\BIBentryALTinterwordspacing
Y.-H. Tseng, L.-H. Lee, L.-P. Chang, and H.-H. Chen, ``Introduction to {SIGHAN} 2015 bake-off for {C}hinese spelling check,'' in \emph{Proceedings of the Eighth {SIGHAN} Workshop on {C}hinese Language Processing}.\hskip 1em plus 0.5em minus 0.4em\relax Beijing, China: Association for Computational Linguistics, Jul. 2015, pp. 32--37. [Online]. Available: \url{https://aclanthology.org/W15-3106}
\BIBentrySTDinterwordspacing

\bibitem{xu-etal-2021-read}
\BIBentryALTinterwordspacing
H.-D. Xu, Z.~Li, Q.~Zhou, C.~Li, Z.~Wang, Y.~Cao, H.~Huang, and X.-L. Mao, ``Read, listen, and see: Leveraging multimodal information helps {C}hinese spell checking,'' in \emph{Findings of the Association for Computational Linguistics: ACL-IJCNLP 2021}.\hskip 1em plus 0.5em minus 0.4em\relax Online: Association for Computational Linguistics, Aug. 2021, pp. 716--728. [Online]. Available: \url{https://aclanthology.org/2021.findings-acl.64}
\BIBentrySTDinterwordspacing

\bibitem{xu-etal-2022-fcgec}
\BIBentryALTinterwordspacing
L.~Xu, J.~Wu, J.~Peng, J.~Fu, and M.~Cai, ``{FCGEC}: Fine-grained corpus for {C}hinese grammatical error correction,'' in \emph{Findings of the Association for Computational Linguistics: EMNLP 2022}.\hskip 1em plus 0.5em minus 0.4em\relax Abu Dhabi, United Arab Emirates: Association for Computational Linguistics, Dec. 2022, pp. 1900--1918. [Online]. Available: \url{https://aclanthology.org/2022.findings-emnlp.137}
\BIBentrySTDinterwordspacing

\bibitem{DBLP:journals/corr/abs-2305-13225}
\BIBentryALTinterwordspacing
Y.~Zhang, L.~Cui, D.~Cai, X.~Huang, T.~Fang, and W.~Bi, ``Multi-task instruction tuning of llama for specific scenarios: {A} preliminary study on writing assistance,'' \emph{CoRR}, vol. abs/2305.13225, 2023. [Online]. Available: \url{https://doi.org/10.48550/arXiv.2305.13225}
\BIBentrySTDinterwordspacing

\bibitem{Lv_2023}
\BIBentryALTinterwordspacing
Q.~Lv, Z.~Cao, L.~Geng, C.~Ai, X.~Yan, and G.~Fu, ``General and domain-adaptive chinese spelling check with error-consistent pretraining,'' \emph{{ACM} Transactions on Asian and Low-Resource Language Information Processing}, vol.~22, no.~5, pp. 1--18, may 2023. [Online]. Available: \url{https://doi.org/10.1145%2F3564271}
\BIBentrySTDinterwordspacing

\bibitem{10.1145/3511808.3557636}
\BIBentryALTinterwordspacing
W.~Jiang, Z.~Ye, Z.~Ou, R.~Zhao, J.~Zheng, Y.~Liu, B.~Liu, S.~Li, Y.~Yang, and Y.~Zheng, ``Mcscset: A specialist-annotated dataset for medical-domain chinese spelling correction,'' in \emph{Proceedings of the 31st ACM International Conference on Information \& Knowledge Management}, ser. CIKM '22.\hskip 1em plus 0.5em minus 0.4em\relax New York, NY, USA: Association for Computing Machinery, 2022, p. 4084–4088. [Online]. Available: \url{https://doi.org/10.1145/3511808.3557636}
\BIBentrySTDinterwordspacing

\bibitem{10.1007/978-3-319-99501-4_41}
Y.~Zhao, N.~Jiang, W.~Sun, and X.~Wan, ``Overview of the nlpcc 2018 shared task: Grammatical error correction,'' in \emph{Natural Language Processing and Chinese Computing}, M.~Zhang, V.~Ng, D.~Zhao, S.~Li, and H.~Zan, Eds.\hskip 1em plus 0.5em minus 0.4em\relax Cham: Springer International Publishing, 2018, pp. 439--445.

\bibitem{zhang-etal-2022-mucgec}
\BIBentryALTinterwordspacing
Y.~Zhang, Z.~Li, Z.~Bao, J.~Li, B.~Zhang, C.~Li, F.~Huang, and M.~Zhang, ``{M}u{CGEC}: a multi-reference multi-source evaluation dataset for {C}hinese grammatical error correction,'' in \emph{Proceedings of the 2022 Conference of the North American Chapter of the Association for Computational Linguistics: Human Language Technologies}.\hskip 1em plus 0.5em minus 0.4em\relax Seattle, United States: Association for Computational Linguistics, Jul. 2022, pp. 3118--3130. [Online]. Available: \url{https://aclanthology.org/2022.naacl-main.227}
\BIBentrySTDinterwordspacing

\bibitem{ma-etal-2022-linguistic}
\BIBentryALTinterwordspacing
S.~Ma, Y.~Li, R.~Sun, Q.~Zhou, S.~Huang, D.~Zhang, L.~Yangning, R.~Liu, Z.~Li, Y.~Cao, H.~Zheng, and Y.~Shen, ``Linguistic rules-based corpus generation for native {C}hinese grammatical error correction,'' in \emph{Findings of the Association for Computational Linguistics: EMNLP 2022}.\hskip 1em plus 0.5em minus 0.4em\relax Abu Dhabi, United Arab Emirates: Association for Computational Linguistics, Dec. 2022, pp. 576--589. [Online]. Available: \url{https://aclanthology.org/2022.findings-emnlp.40}
\BIBentrySTDinterwordspacing

\bibitem{devlin-etal-2019-bert}
\BIBentryALTinterwordspacing
J.~Devlin, M.-W. Chang, K.~Lee, and K.~Toutanova, ``{BERT}: Pre-training of deep bidirectional transformers for language understanding,'' in \emph{Proceedings of the 2019 Conference of the North {A}merican Chapter of the Association for Computational Linguistics: Human Language Technologies, Volume 1 (Long and Short Papers)}.\hskip 1em plus 0.5em minus 0.4em\relax Minneapolis, Minnesota: Association for Computational Linguistics, Jun. 2019, pp. 4171--4186. [Online]. Available: \url{https://aclanthology.org/N19-1423}
\BIBentrySTDinterwordspacing

\bibitem{zhang-etal-2020-spelling}
\BIBentryALTinterwordspacing
S.~Zhang, H.~Huang, J.~Liu, and H.~Li, ``Spelling error correction with soft-masked {BERT},'' in \emph{Proceedings of the 58th Annual Meeting of the Association for Computational Linguistics}.\hskip 1em plus 0.5em minus 0.4em\relax Online: Association for Computational Linguistics, Jul. 2020, pp. 882--890. [Online]. Available: \url{https://aclanthology.org/2020.acl-main.82}
\BIBentrySTDinterwordspacing

\bibitem{li-etal-2021-exploration}
\BIBentryALTinterwordspacing
C.~Li, C.~Zhang, X.~Zheng, and X.~Huang, ``Exploration and exploitation: Two ways to improve {C}hinese spelling correction models,'' in \emph{Proceedings of the 59th Annual Meeting of the Association for Computational Linguistics (Volume 2: Short Papers)}.\hskip 1em plus 0.5em minus 0.4em\relax Online: Association for Computational Linguistics, Aug. 2021, pp. 441--446. [Online]. Available: \url{https://aclanthology.org/2021.acl-short.56}
\BIBentrySTDinterwordspacing

\bibitem{li-etal-2022-learning-dictionary}
\BIBentryALTinterwordspacing
Y.~Li, S.~Ma, Q.~Zhou, Z.~Li, L.~Yangning, S.~Huang, R.~Liu, C.~Li, Y.~Cao, and H.~Zheng, ``Learning from the dictionary: Heterogeneous knowledge guided fine-tuning for {C}hinese spell checking,'' in \emph{Findings of the Association for Computational Linguistics: EMNLP 2022}.\hskip 1em plus 0.5em minus 0.4em\relax Abu Dhabi, United Arab Emirates: Association for Computational Linguistics, Dec. 2022, pp. 238--249. [Online]. Available: \url{https://aclanthology.org/2022.findings-emnlp.18}
\BIBentrySTDinterwordspacing

\bibitem{shao2021cpt}
Y.~Shao, Z.~Geng, Y.~Liu, J.~Dai, F.~Yang, L.~Zhe, H.~Bao, and X.~Qiu, ``Cpt: A pre-trained unbalanced transformer for both chinese language understanding and generation,'' \emph{arXiv preprint arXiv:2109.05729}, 2021.

\bibitem{lewis2020bart}
M.~Lewis, Y.~Liu, N.~Goyal, M.~Ghazvininejad, A.~Mohamed, O.~Levy, V.~Stoyanov, and L.~Zettlemoyer, ``Bart: Denoising sequence-to-sequence pre-training for natural language generation, translation, and comprehension,'' in \emph{Proceedings of the 58th Annual Meeting of the Association for Computational Linguistics}, 2020, pp. 7871--7880.

\bibitem{dong2022survey}
C.~Dong, Y.~Li, H.~Gong, M.~Chen, J.~Li, Y.~Shen, and M.~Yang, ``A survey of natural language generation,'' \emph{ACM Computing Surveys}, vol.~55, no.~8, pp. 1--38, 2022.

\bibitem{zhang2022mucgec}
Y.~Zhang, Z.~Li, Z.~Bao, J.~Li, B.~Zhang, C.~Li, F.~Huang, and M.~Zhang, ``Mucgec: a multi-reference multi-source evaluation dataset for chinese grammatical error correction,'' in \emph{Proceedings of the 2022 Conference of the North American Chapter of the Association for Computational Linguistics: Human Language Technologies}, 2022, pp. 3118--3130.

\bibitem{wei2020structbert}
\BIBentryALTinterwordspacing
W.~Wang, B.~Bi, M.~Yan, C.~Wu, J.~Xia, Z.~Bao, L.~Peng, and L.~Si, ``Structbert: Incorporating language structures into pre-training for deep language understanding,'' in \emph{8th International Conference on Learning Representations, {ICLR} 2020, Addis Ababa, Ethiopia, April 26-30, 2020}.\hskip 1em plus 0.5em minus 0.4em\relax OpenReview.net, 2020. [Online]. Available: \url{https://openreview.net/forum?id=BJgQ4lSFPH}
\BIBentrySTDinterwordspacing

\bibitem{omelianchuk2020gector}
K.~Omelianchuk, V.~Atrasevych, A.~Chernodub, and O.~Skurzhanskyi, ``Gector--grammatical error correction: Tag, not rewrite,'' in \emph{Proceedings of the Fifteenth Workshop on Innovative Use of NLP for Building Educational Applications}, 2020, pp. 163--170.

\bibitem{vicuna2023}
\BIBentryALTinterwordspacing
W.-L. Chiang, Z.~Li, Z.~Lin, Y.~Sheng, Z.~Wu, H.~Zhang, L.~Zheng, S.~Zhuang, Y.~Zhuang, J.~E. Gonzalez, I.~Stoica, and E.~P. Xing, ``Vicuna: An open-source chatbot impressing gpt-4 with 90\%* chatgpt quality,'' March 2023. [Online]. Available: \url{https://lmsys.org/blog/2023-03-30-vicuna/}
\BIBentrySTDinterwordspacing

\bibitem{du2022glm}
Z.~Du, Y.~Qian, X.~Liu, M.~Ding, J.~Qiu, Z.~Yang, and J.~Tang, ``Glm: General language model pretraining with autoregressive blank infilling,'' in \emph{Proceedings of the 60th Annual Meeting of the Association for Computational Linguistics (Volume 1: Long Papers)}, 2022, pp. 320--335.

\bibitem{ye2023cleme}
J.~Ye, Y.~Li, Q.~Zhou, Y.~Li, S.~Ma, H.-T. Zheng, and Y.~Shen, ``Cleme: Debiasing multi-reference evaluation for grammatical error correction,'' \emph{arXiv preprint arXiv:2305.10819}, 2023.

\bibitem{zhang2009features}
B.~Zhang, ``Features and functions of the hsk dynamic composition corpus,'' \emph{International Chinese Language Education}, vol.~4, pp. 71--79, 2009.

\bibitem{rao-etal-2018-overview}
\BIBentryALTinterwordspacing
G.~Rao, Q.~Gong, B.~Zhang, and E.~Xun, ``Overview of {NLPTEA}-2018 share task {C}hinese grammatical error diagnosis,'' in \emph{Proceedings of the 5th Workshop on Natural Language Processing Techniques for Educational Applications}.\hskip 1em plus 0.5em minus 0.4em\relax Melbourne, Australia: Association for Computational Linguistics, Jul. 2018, pp. 42--51. [Online]. Available: \url{https://aclanthology.org/W18-3706}
\BIBentrySTDinterwordspacing

\bibitem{rao-etal-2020-overview}
\BIBentryALTinterwordspacing
G.~Rao, E.~Yang, and B.~Zhang, ``Overview of {NLPTEA}-2020 shared task for {C}hinese grammatical error diagnosis,'' in \emph{Proceedings of the 6th Workshop on Natural Language Processing Techniques for Educational Applications}.\hskip 1em plus 0.5em minus 0.4em\relax Suzhou, China: Association for Computational Linguistics, Dec. 2020, pp. 25--35. [Online]. Available: \url{https://aclanthology.org/2020.nlptea-1.4}
\BIBentrySTDinterwordspacing

\bibitem{wang2021yaclc}
Y.~Wang, C.~Kong, L.~Yang, Y.~Wang, X.~Lu, R.~Hu, S.~He, Z.~Liu, Y.~Chen, E.~Yang, and M.~Sun, ``Yaclc: A chinese learner corpus with multidimensional annotation,'' 2021.

\bibitem{wu2023chatgpt}
H.~Wu, W.~Wang, Y.~Wan, W.~Jiao, and M.~Lyu, ``Chatgpt or grammarly? evaluating chatgpt on grammatical error correction benchmark,'' 2023.

\bibitem{fang2023chatgpt}
T.~Fang, S.~Yang, K.~Lan, D.~F. Wong, J.~Hu, L.~S. Chao, and Y.~Zhang, ``Is chatgpt a highly fluent grammatical error correction system? a comprehensive evaluation,'' 2023.

\bibitem{coyne2023analyzing}
S.~Coyne, K.~Sakaguchi, D.~Galvan-Sosa, M.~Zock, and K.~Inui, ``Analyzing the performance of gpt-3.5 and gpt-4 in grammatical error correction,'' 2023.

\end{thebibliography}

\begin{CJK*}{UTF8}{gbsn}
\begin{IEEEbiography}[{\includegraphics[width=1in,height=1.25in,clip,keepaspectratio]{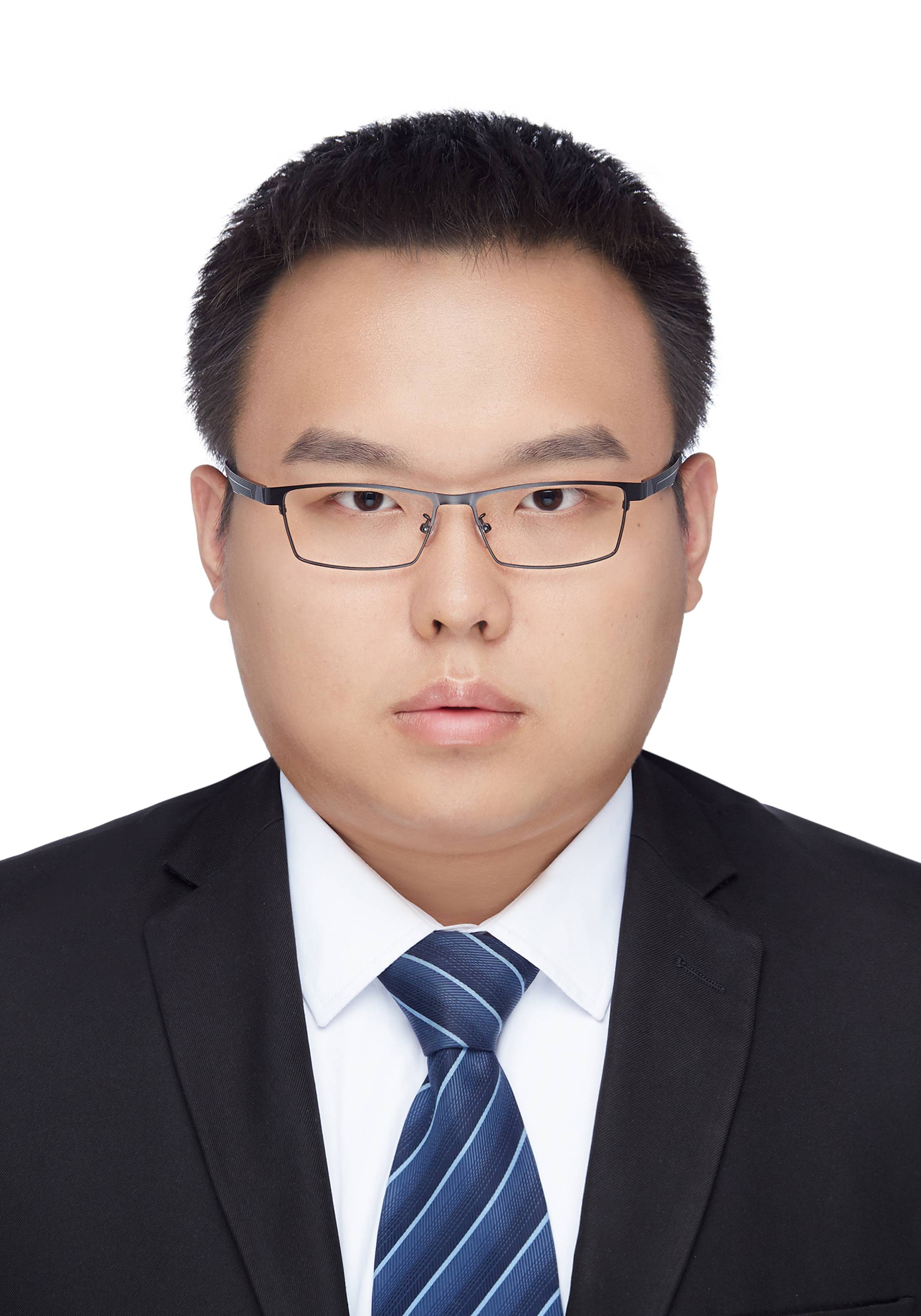}}]{Yinghui Li} received the BEng degree from the Department of Computer Science and Technology, Tsinghua University, in 2020. He is currently working toward the PhD degree with the Tsinghua Shenzhen International Graduate School, Tsinghua University. His research interests include natural language processing and deep learning.
\end{IEEEbiography}

\begin{IEEEbiography}[{\includegraphics[width=1in,height=1.25in,clip,keepaspectratio]{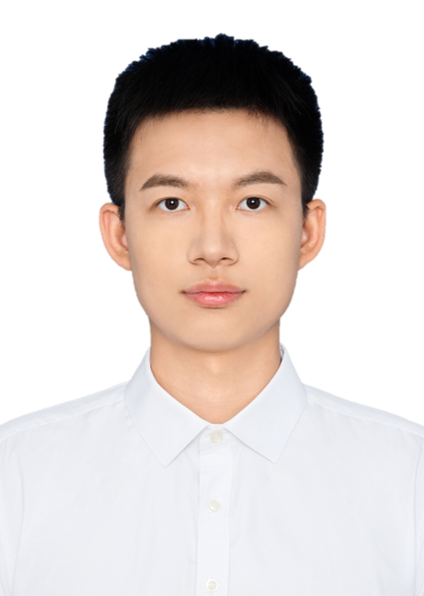}}]{Haojing Huang} received the BEng degree from College of Mathematics and Information \& College of Software Engineering, South China Agriculture University, in 2023. He is currently working toward a Master's degree with the Tsinghua Shenzhen International Graduate School, Tsinghua University. His research interests include natural language processing and deep learning.
\end{IEEEbiography}

\begin{IEEEbiography}[{\includegraphics[width=1in,height=1.25in,clip,keepaspectratio]{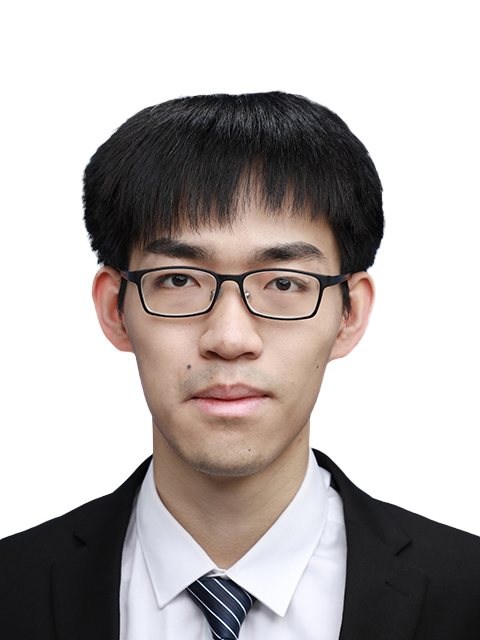}}]{Shirong Ma} received the BEng degree from the Department of Computer Science and Technology, Tsinghua University, in 2021. He is currently working toward the Master degree with the Tsinghua Shenzhen International Graduate School, Tsinghua University. His research interests include natural language processing and deep learning.
\end{IEEEbiography}

\begin{IEEEbiography}[{\includegraphics[width=1in,height=1.25in,clip,keepaspectratio]{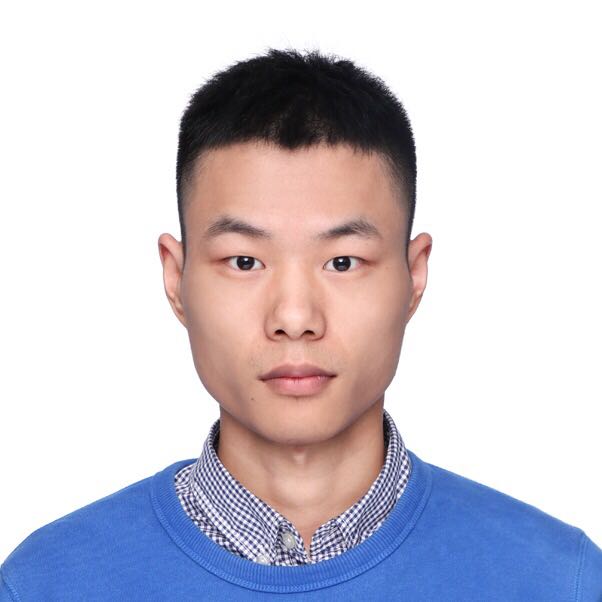}}]{Yong Jiang} is a researcher at DAMO Academy, Alibaba Group. He leads a research team tackling complex problems related to machine learning and NLP. He received his PhD degree from the joint program of ShanghaiTech University and University of Chinese Academy of Sciences. He was a visiting scholar at University of California, Berkeley. He has over 50 publications in top-tier conferences and journals including ACL, EMNLP, NAACL, AAAI, and IJCAI. He has received the best system paper awards from SemEval 2022 and 2023. 
\end{IEEEbiography}

\begin{IEEEbiography}[{\includegraphics[width=1in,height=1.25in,clip,keepaspectratio]{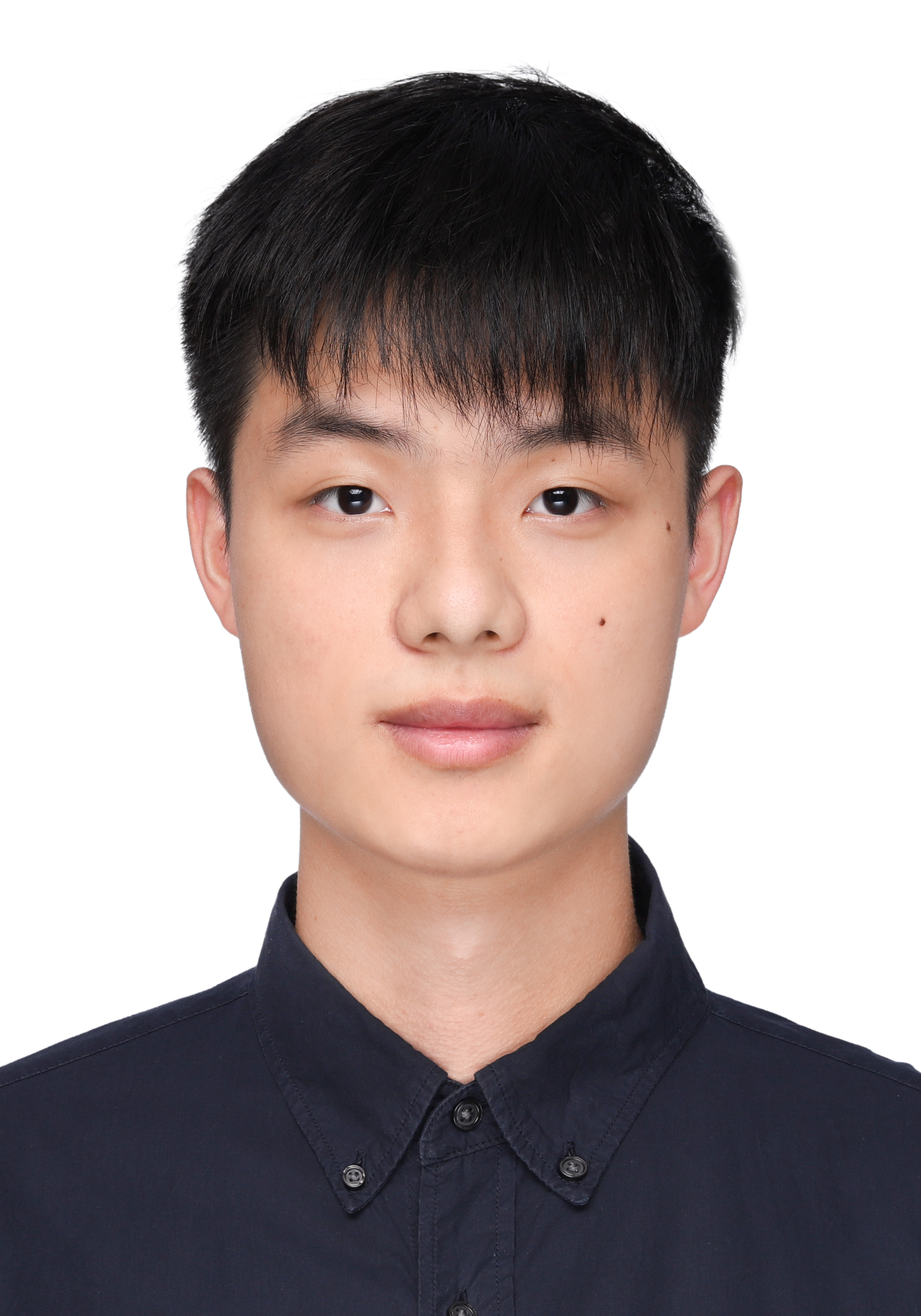}}]{Yangning Li} received the BEng degree from the Department of Computer Science and Technology, Huazhong University of Science and Technology, in 2020. He is currently working toward a Master's degree with the Tsinghua Shenzhen International Graduate School, Tsinghua University. His research interests include natural language processing and data mining.
\end{IEEEbiography}

\begin{IEEEbiography}[{\includegraphics[width=1in,height=1.25in,clip,keepaspectratio]{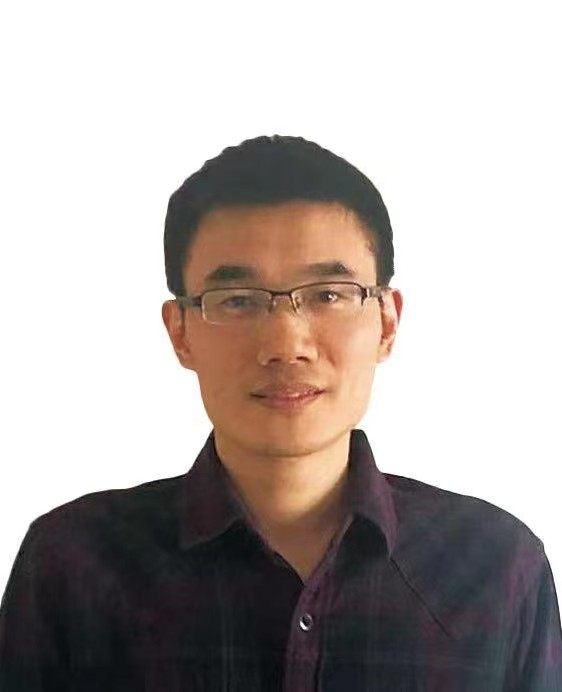}}]{Feng Zhou} received the M.S. degree computer science from Beijing University of Posts and Telecommunications, Beijing, China. He is currently the technical expert and leader of the Tensor Lab of OPPO Research Institute. His research interests include Intelligent Dialogue System, Natural Language Processing, Pre-training and Application of LLM.
\end{IEEEbiography}

\begin{IEEEbiography}[{\includegraphics[width=1in,height=1.25in,clip,keepaspectratio]{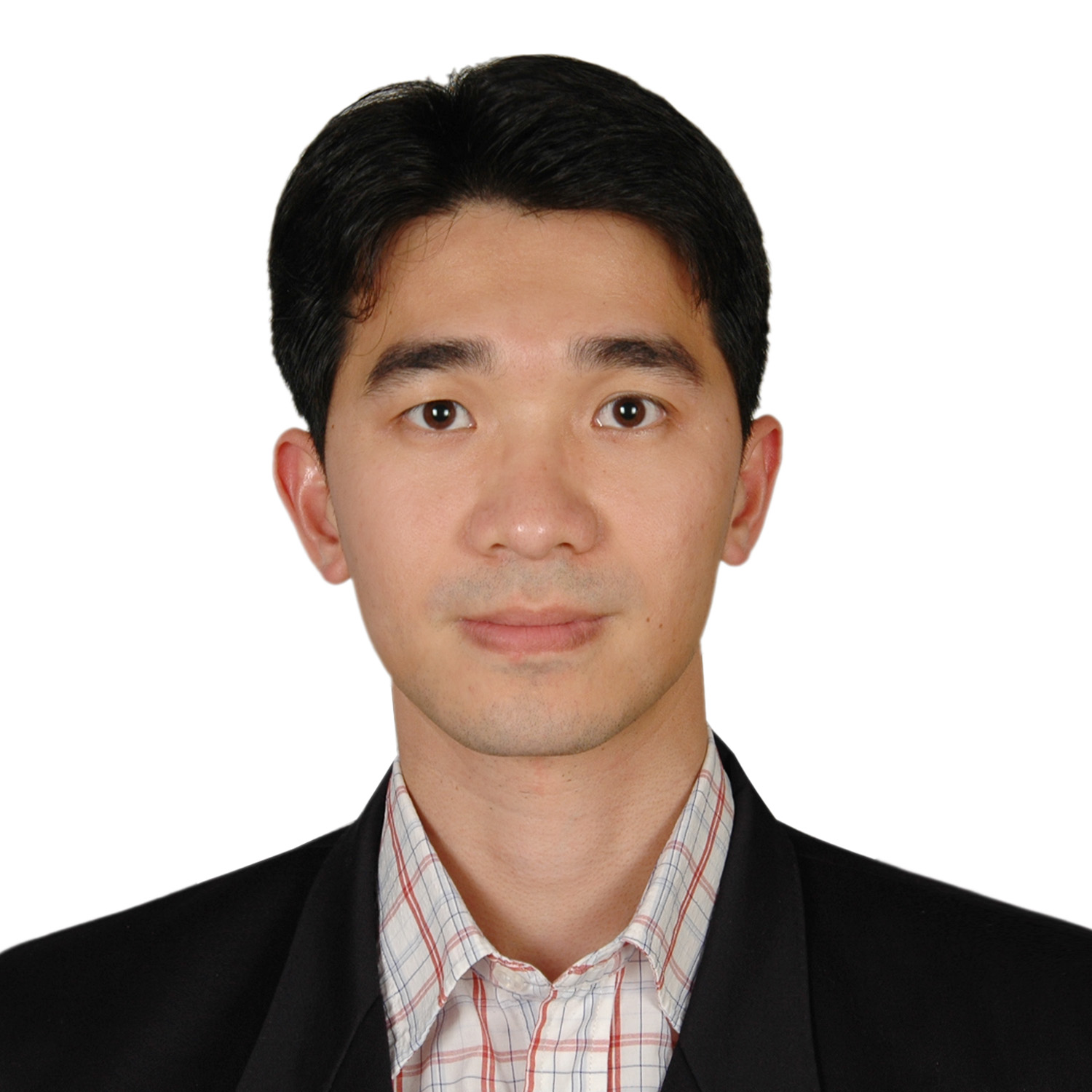}}]{Hai-Tao Zheng} received the PhD degree in medical informatics from Seoul National University, South Korea. He is currently an associate professor with the Shenzhen International Graduate School, Tsinghua University, and also with Peng Cheng Laboratory. His research interests include web science, semantic web, information retrieval, and machine learning. He has authored or coauthored 80+ papers in top international academic conferences and journals, such as Natural Machine Intelligence, IEEE TKDE, ICLR, AAAI, IJCAI, SIGIR, ACL, EMNLP, and ACM MM. He was the Area Chair of NeurIPS 2023, Session Chair of AAAI 2023, Program Committee Member of AAAI, ACL, SIGKDD, MM, and Reviewer of TKDE and TASLP. The OpenPrompt he proposed has received the Best Demo Paper Award from ACL 2022.
\end{IEEEbiography}

\begin{IEEEbiography}[{\includegraphics[width=1in,height=1.25in,clip,keepaspectratio]{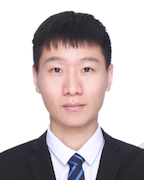}}]{Qingyu Zhou} received his PhD degree in a joint program of Harbin Institute of Technology and Microsoft Research Asia. He is currently a Senior Researcher at the Tensor Lab of OPPO Research Institute. His research interests include text summarization, natural language generation, dialogue system, and educational applications of NLP.
\end{IEEEbiography}
\end{CJK*}

% \vfill
\end{CJK*}
\end{document}